\documentclass[11pt, a4paper]{article}
\usepackage[utf8]{inputenc}
\usepackage{geometry}
\usepackage{graphicx}
\usepackage{hyperref}
\usepackage{cite}
\usepackage{array}
\usepackage{booktabs} 
\usepackage{amsmath}
\usepackage{amssymb}
\usepackage{tikz}
\usetikzlibrary{arrows.meta, positioning, calc, fit, backgrounds}
\usepackage{tikz}
\usepackage{amsmath}
\usepackage{graphicx}
\usetikzlibrary{arrows.meta,calc,fit,positioning}
\usepackage{soul}
\usepackage{xcolor}
\sethlcolor{yellow!50}
\usepackage{subcaption}
\usepackage{makecell}
\usepackage{multirow}
\usepackage{rotating}
\usepackage{algorithm}
\usepackage{algpseudocode}
\usepackage{float}

\title{CARMA: Correlation-Aware Reconstruction for Multivariate Time Series Anomaly Detection}
\author{
Md. Johir Raihan$^{1}$ \quad
Zahra Zamanzadeh Darban$^{1}$ \quad
Yueyang Liu$^{1}$ \\[0.3em]
Flora D. Salim$^{2}$ \quad
Levin Kuhlmann$^{1}$ \quad
Mahsa Salehi$^{1}$ \\[0.4em]
{\small $^{1}$ Monash University, Melbourne, Victoria, Australia} \\
{\small $^{2}$ University of New South Wales, Sydney, New South Wales, Australia}
}
\date{}

\begin{document}

\maketitle

\section*{Abstract}
Reconstruction-based detectors for multivariate time series anomaly detection typically learn a single entangled representation of normal behavior and flag anomalies based on reconstruction error. However, real-world anomalies are not only isolated spikes in a single channel but also disruptions in the relationships between channels, where each sensor individually appears normal, but their joint behavior deviates from the norm. Because these models compress both temporal evolution and inter-channel structure into this single space, they lack an explicit mechanism to isolate such inter-channel dependencies from normal temporal dependencies, leading to lower detection rates for cross-channel anomalies. We propose CARMA (Correlation-Aware Reconstruction for Multivariate Time Series Anomaly Detection), which is a reconstruction-based variational autoencoder with two complementary latent spaces: a temporal latent that captures sequential dynamics, and a correlation latent that is trained to reconstruct the window's pairwise Pearson correlations. Both representations are fused in the signal decoder, and anomalies are detected from the reconstruction error. We evaluate CARMA on the TSB-AD-M benchmark suite (17 datasets, 180 series) against 29 baseline methods, and show that it significantly outperforms all of them across multiple threshold-free and range-aware metrics, achieving a VUS-PR of 0.4929. Ablations show that removing the correlation branch degrades performance, that adding a second latent without supervision recovers only part of the gain, and that supervising the correlation latent with an explicit correlation target accounts for the rest and produces a sharper error response at anomalies. URL: \url{anonymous.4open.science/r/CARMA-AE8F}


\section{Introduction}
Time series Anomaly Detection (TSAD) in multivariate streams is a core problem in monitoring cyber-physical systems, industrial sensors, medical recordings, and other multi-channel streams \cite{darbanDeepLearningTime2025, liu2024Elephant}. In these settings, anomalies are not solely characterized by large deviations on individual channels, but also by changes in the inter-channel structure \cite{sunLatentFlowDiscoveringLatent2026}. Specifically, correlation breaks, or phase mismatches, can induce mildly abnormal behavior in individual trajectories. A detector that reasons only about per-channel amplitude can therefore miss events that are obvious once inter-channel structure is taken into account.

\textbf{Motivation.} Reconstruction-based detectors \cite{darbanDeepLearningTime2025}, such as OmniAnomaly \cite{suRobustAnomalyDetection2019}, xLSTMAD~\cite{faberXLSTMADPowerfulXLSTMbased2025}, and Stream-VAE~\cite{ozer2026STREAMVAE}, learn to reconstruct normal multivariate dynamics and use deviations between the input and its reconstruction to detect anomalies. However, because the reconstruction loss only measures per-channel error, flexible decoders can reproduce broken inter-channel dependencies, and as a result they achieve a low reconstruction error even if the relationships between channels are not normal. Furthermore, without explicitly modeling these intrinsic inter-channel correlations, the model may fail to accurately reconstruct even entirely normal time windows, thereby introducing errors and compromising overall detection reliability.

\begin{figure}[t]
    \centering
    \includegraphics[width=0.8\linewidth]{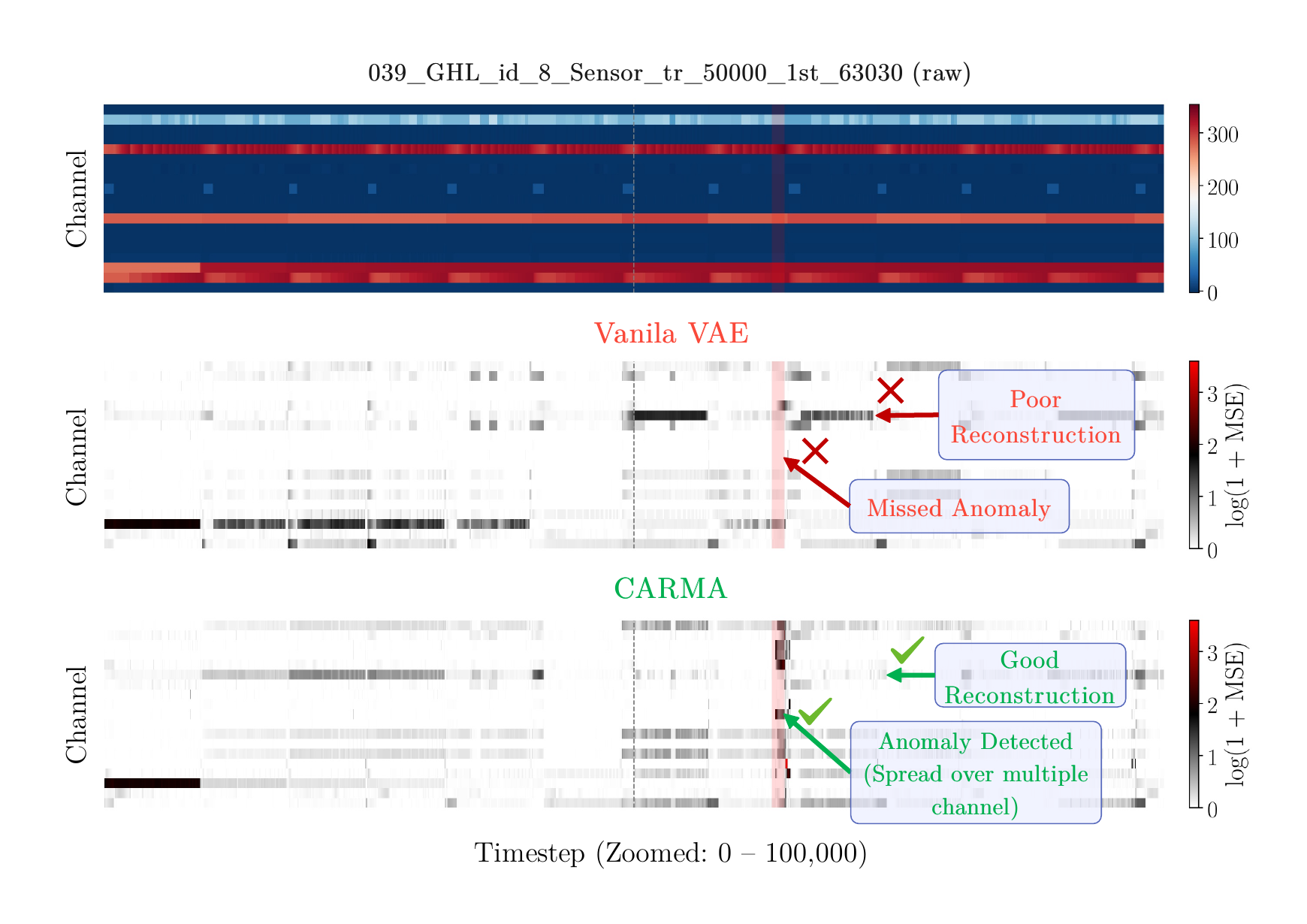}
\caption{Top panel shows GHL sensor data with the ground truth anomaly shaded. Middle panel displays $\log(1+\mathrm{MSE})$ for CARMA without correlation conditioning, which misses the anomaly with spurious errors in normal regions. Bottom panel shows full CARMA achieving low reconstruction during normal segments with localized multichannel error increase at the anomalous interval.}
    \label{fig:intro}
\end{figure}

This is not hypothetical. Figure~\ref{fig:intro} illustrates this issue on a GHL sensor series\cite{GHL2016URL, filonovMultivariateIndustrialTime2016}. Without correlation conditioning (Vanilla VAE, middle panel), the model fails to properly reconstruct normal operating windows ($\textit{Poor Reconstruction}$), producing spurious reconstruction errors that trigger false alarms. Concurrently, during a true anomaly event, its decoder improperly adapts to the abnormal channel dynamics, leading to low reconstruction error and a missed detection ($\textit{Missed Anomaly}$). In contrast, CARMA (bottom panel) maintains low reconstruction error during normal operation and exhibits a sharp multichannel error increase aligned with the ground-truth boundary. The contrast shows that explicit dependence modelling changes how reconstruction error is distributed in time and across channels, making the labeled event easier to separate from normal behaviour.

Our approach extends standard signal reconstruction by treating inter-channel dependence as a generative component rather than an implicit byproduct. 
We instantiate this as \textbf{CARMA} (Correlation-Aware Reconstruction for Multivariate time series Anomalies), a variational autoencoder (VAE) with two complementary latents. From each sliding window, the encoder forms two representations. A temporal latent $z_t$ is inferred at every timestep and encodes local dynamics, while a single window-level latent $z_c$ encodes how channels correlate over that window. A head on $z_c$ predicts the window correlation pattern and is trained against correlations computed from the input. The signal decoder combines $z_t$ with $z_c$ at each timestep so reconstruction uses both temporal dynamics and cross-channel structure. This architectural coupling makes every predicted timestep explicitly conditional on the model’s estimate of inter-channel dependence. 
In this design, correlation modelling is not an auxiliary regularizer, rather it is a second generative factor that the model must resolve, and that the signal reconstruction is allowed to use.
The training objective jointly minimises signal reconstruction error, correlation reconstruction error, and KL divergence, but at test time we score anomalies solely from the signal reconstruction error. Because \(z_{\mathrm{c}}\) shapes what the decoder considers reconstructible under normal dependence, structural deviations produce elevated signal error even when marginal values remain within expected ranges.

We evaluate CARMA on the multivariate track of TSB-AD \cite{liu2024Elephant}, a large and heterogeneous benchmark spanning 17 datasets and 180 series. Our contributions are as follows:


\begin{enumerate}
    \item We identify that the known tendency of reconstruction models to reconstruct anomalies too well \cite{gongMemorizingNormalityDetect2019} takes a specific form in multivariate time series data: because the reconstruction loss only measures per-channel error, a flexible decoder can reproduce broken inter-channel dependence at low error, while the lack of explicit correlation modelling causes poor reconstruction of normal windows and false alarms.
    \item We propose CARMA, a VAE with a temporal latent $z_{\mathrm{t}}$ and an inter-channel latent $z_{\mathrm{c}}$. Unlike prior dual-latent designs, $z_{\mathrm{c}}$ is supervised to reconstruct an explicit Pearson correlation summary and is injected into every timestep of the signal decoder.
    \item We show that the gain comes from supervising $z_{\mathrm{c}}$ with an explicit correlation target rather than from added latent capacity, and that this supervision produces a sharper reconstruction-error response at anomalies.
    \item Our extensive empirical results on the TSB-AD-M benchmark with 17 datasets (180 series) show that CARMA \textit{significantly} outperforms 29 baselines. We further compare CARMA to an ablated model that is identical except for the correlation branch, and show that the full model achieves significantly higher per-series scores in a paired test over all series.

\end{enumerate}

\section{Related Work}
A large number of Multivariate Time Series Anomaly Detection (MTSAD) methods learn behavior from normal windows and flags deviations at test time, yet their design prioritizes temporal structure over explicit modeling of inter-channel relationships. 

\paragraph{Implicit channel mixing in temporal architectures.} Anomaly Transformer embeds each multivariate timestep into a shared latent sequence and applies Anomaly-Attention, which compares a learned series-association map to a prior over temporal lags \cite{xu2022Anomaly}. The objective emphasizes temporal association patterns rather than an explicit estimate of inter-channel correlation within each window, and it does not condition decoding on a dedicated channel-dependence representation. CARLA~\cite{darban2025CARLA} uses a multi-channel ResNet to learn window embeddings under contrastive and self-supervised objectives, whereas COUTA~\cite{xu2024Calibrated} transforms the channel-first TCN features into a one-class representation centred on a hypersphere. NCAD compares the context and target segments by means of a shared TCN encoder and a contextual hypersphere classifier \cite{carmona2022Neural}. In each of these cases, although the multivariate signals are processed jointly, none of them establishes an explicit pairwise channel-dependence target to guide normal signal modeling and decoding.

\textbf{Explicit correlation and dependency modeling.}
A smaller set of methods tries to directly model the relationships between variables, rather than relying on implicit mixing. MTAD-GAT \cite{zhao2020Multivariate} applies graph attention over both the feature and temporal dimensions, treating channels as a fully connected graph, and jointly optimises forecasting and reconstruction objectives. GDN \cite{deng2021Graph} learns a static sensor graph and uses graph attention to forecast each channel. CST-GL \cite{zhengCorrelationAwareSpatialTemporal2024} learns a directed sensor graph and uses it for message passing in a forecasting Spatio-temporal GNN, scoring anomalies from forecast residuals. In these models, inter-channel structure is used as an inductive bias for predicting or reconstructing per-channel values. 
Other work models dependence through time-varying graphs (DyGraphAD \cite{chenMultivariateTimeSeries2023}), graph alignment (MADGA \cite{wangInterdependencyMattersGraph2024}), Granger causality (GCAD \cite{liu2025GCAD}), or learned channel masks in the frequency domain (CATCH \cite{wuCATCHChannelAwareMultivariate2026}). In all of these, dependence guides prediction or fusion rather than serving as an explicit reconstruction target.
InterFusion \cite{liMultivariateTimeSeries2021a} uses a hierarchical VAE with separate inter-channel and temporal latents, but the inter-channel latent is learned without an explicit dependency target. Our ablation in Sec.~\ref{sec:abl} isolates the effect of this supervision.
Closest to our work, MSCRED \cite{chuxuzhangDeepNeuralNetwork} reconstructs matrices of inter-channel correlation and scores anomalies from their residuals.
None of these methods uses a dedicated window-level correlation latent that is used to predict an explicit pairwise correlation summary and conditions the reconstruction of the signal itself.
CARMA instead learns separate temporal and correlation representations of normal multivariate windows, so channel-dependency structure is modelled as its own component rather than as an indirect effect of attention or convolution. This makes CARMA well suited to anomalies that disrupt normal channel relationships without large per-channel amplitude deviations.

\section{Methodology}
\subsection{Problem Statement}
Let $\mathcal{D}=\{X_i\}_{i=1}^{N}$ denote a training set of multivariate time series windows, where each window $X_i \in \mathbb{R}^{T \times D}$ contains $T$ time steps and $D$ variables. In the unsupervised anomaly detection setting, training windows are assumed to be predominantly normal.
We learn a probabilistic reconstruction model that maps an input window $X$ to both a series reconstruction $\hat{X}$ and a correlation reconstruction $\hat{r}$, using training data. At test time, anomalies are detected from the reconstruction loss,
\begin{equation}
    s(X) = \mathcal{L}_{\text{rec}}(X),
    \label{eq:anomaly_score}
\end{equation}
where larger scores indicate stronger deviation from the learned normal behavior. The goal is to learn a reconstruction model that captures both temporal dynamics and inter-channel dependency structure, so that large reconstruction error reflects deviations in channel relationships as well as temporal patterns.

\subsection{CARMA Method}
\begin{figure*}[htb]
    \centering
    \includegraphics[width=0.95\linewidth]{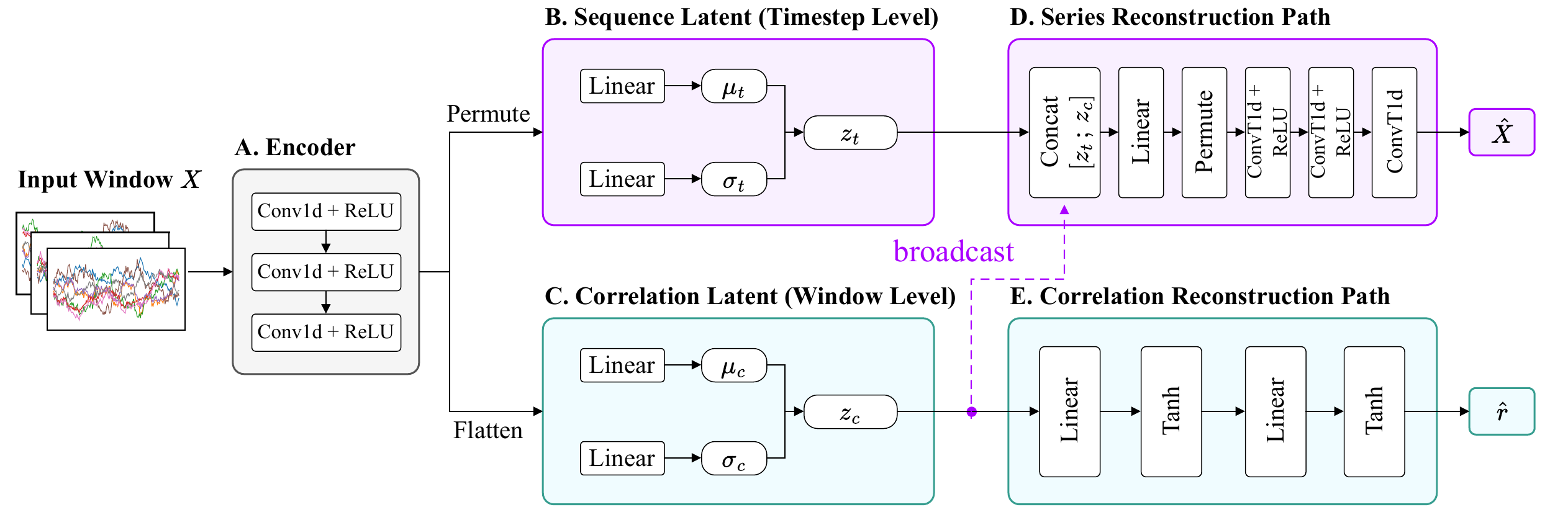}
    \caption{Overview of CARMA. A multivariate window is encoded into temporal and correlation latents. The concatenated latents reconstruct the input series, while the correlation latent reconstructs window-level channel dependencies.}
    \label{fig:architecture}
\end{figure*}

\textbf{CARMA Overview.} Our proposed CARMA framework is illustrated in Figure \ref{fig:architecture}. Each input window is passed through a shared temporal encoder that produces a feature map over time and channels. One branch keeps that map at timestep resolution and infers a temporal latent $z_t$ at every step, giving a sequence of codes for how the multivariate signal evolves within the window. The other branch flattens the map across time and infers one window-level correlation latent $z_c$ that summarizes dependence among channels. For series reconstruction, $z_c$ is broadcast to each timestep and concatenated with the temporal latent at that step before the series decoder outputs $\hat{X}$. The same $z_c$ feeds a separate correlation decoder that predicts the within-window dependency vector $\hat{r}$. Supervising both $\hat{X}$ and $\hat{r}$ encourages the model to retain measurable cross-channel structure instead of hiding it in a single series bottleneck. CARMA thus models step-level dynamics and window-level correlations via factorized pathways while recombining them for final series reconstruction.

\noindent
\textbf{Encoder.}
Given an input window $X \in \mathbb{R}^{T \times D}$, we transpose it to channel-first form $X^{\top} \in \mathbb{R}^{D \times T}$ and pass it through a shared temporal encoder $f_{\phi}$ composed of three Conv1d--ReLU layers. The first layer uses all $D$ variables as input channels, so channels are mixed while convolution runs along the time axis. The encoder returns a feature map $H = f_{\phi}(X^{\top}) \in \mathbb{R}^{d_e \times T}$ with the same length $T$ as the input window. Two readouts branch from $H$. For the temporal path, we rearrange  encoder output $H$ to timestep-major form. Then each timestep $\tau$ is associated with its own feature vector $h_\tau \in \mathbb{R}^{d_e}$, which contains the local multivariate pattern at that point in the window. The same linear inference heads are applied at every $\tau$, producing step-level variational parameters for the temporal latent. While the temporal branch reads $H$ one timestep at a time, we summarize cross-channel structure over the full window on the correlation branch. We flatten the encoder map into $h_{\mathrm{flat}} = \mathrm{vec}(H) \in \mathbb{R}^{d_h}$, where $d_h = d_e T$ collects every encoder features of $H$ across all $T$ steps. With temporal and correlation inference maps $f_t(\cdot)$ and $f_c(\cdot)$, the variational parameters can be written as
\begin{equation}
    (\mu_{t,\tau}, \log \sigma_{t,\tau}^2) = f_t(h_{\tau}), \qquad
    (\mu_c, \log \sigma_c^2) = f_c(h_{\mathrm{flat}}),
    \label{eq:encoder_heads}
\end{equation}
which specifies the Gaussian parameters of the step-level temporal latent at each $\tau$ and of the window-level correlation latent $z_c$.

\noindent
\textbf{Latent branches.}
Conditioned on the encoder embeddings, CARMA places a diagonal Gaussian over the correlation latent and, at each timestep, a separate diagonal Gaussian over the temporal latent. The correlation approximate posterior is
\begin{equation}
    q_{\phi}(z_c \mid X) = \mathcal{N}\!\left(z_c \mid \mu_c, \mathrm{diag}(\sigma_c^2)\right),
\end{equation}
with $(\mu_c, \log \sigma_c^2)$ given by $f_c(h_{\mathrm{flat}})$ in~\eqref{eq:encoder_heads}. The temporal latents factorize across timesteps,
\begin{equation}
    q_{\phi}(z_t \mid X) = \prod_{\tau=1}^{T} q_{\phi}(z_{t,\tau} \mid X), \qquad
    q_{\phi}(z_{t,\tau} \mid X) = \mathcal{N}\!\left(z_{t,\tau} \mid \mu_{t,\tau}, \mathrm{diag}(\sigma_{t,\tau}^2)\right),
\end{equation}
where $(\mu_{t,\tau}, \log \sigma_{t,\tau}^2) = f_t(h_{\tau})$. During training, samples are drawn with the reparameterization trick,
\begin{equation}
    z_{t,\tau} = \mu_{t,\tau} + \epsilon_{\tau} \odot \sigma_{t,\tau}, \qquad
    z_c = \mu_c + \epsilon_c \odot \sigma_c,
    \quad \epsilon_{\tau}, \epsilon_c \sim \mathcal{N}(0, I),
\end{equation}
with independent $\epsilon_{\tau}$ at each $\tau$. At inference, we use the posterior means ($z_{t,\tau}=\mu_{t,\tau}$, $z_c=\mu_c$), as in standard VAE practice. The sequence $\{z_{t,\tau}\}_{\tau=1}^{T}$ encodes how the multivariate signal evolves step by step within the window, while $z_c$ encodes window-level dependence among channels and is used for both series and correlation reconstruction.

\noindent
\textbf{Correlation decoder.}
 Methods that share a joint encoder can mix channels in the representation, but they do not require the latent code to preserve a measurable dependency structure. As a result, correlation shifts may remain only weakly reflected in the reconstruction error. To address this, for each window, we first compute a target dependency vector $r(X) \in \mathbb{R}^{D(D-1)/2}$ from the lower-triangular entries of the within-window Pearson correlation matrix. Pearson correlation provides a scale-normalized summary of pairwise channel co-variation within the window, and the lower-triangular vectorization removes redundant symmetric entries while retaining all unique pairs. The correlation latent $z_c$ is mapped by a decoder $h_{\theta}$ into a predicted dependency vector $\hat{r}(X)=h_{\theta}(z_c)$. Concretely, $z_c$ is passed through two linear layers with Tanh activations to produce $\hat{r}(X)$.

\noindent \textbf{Series decoder.} 
For multivariate series reconstruction, the window-level correlation latent $z_c$ is broadcast to each timestep $\tau$ and concatenated with the step-level temporal latent $z_{t,\tau}$, producing a combined code $[z_{t,\tau}; z_c]$ at every position in the window. A separate decoder network $g_{\theta}$ maps this sequence to the reconstructed window $\hat{X}=g_{\theta}\bigl(\{[z_{t,\tau}; z_c]\}_{\tau=1}^{T}\bigr)\in\mathbb{R}^{T\times D}$. Internally, $g_{\theta}$ projects the per-timestep latent vectors back to a multichannel time series through a transposed-convolution decoder that mirrors the temporal width of the encoder (length $T$, $D$ output channels). Conditioning on both $z_{t,\tau}$ and $z_c$ lets the series path recover local dynamics from the temporal latents while tying each reconstructed timestep to the window-level dependence encoded in $z_c$.

\subsection{CARMA's Training Objective}
\noindent
\textbf{Series Reconstruction Loss.}
The series decoder is trained to recover the input window from the latent codes fed into $g_{\theta}$. We use the mean squared error between $X$ and its reconstruction $\hat{X}$, averaged over all timesteps and channels in the window,
\begin{equation}
    \mathcal{L}_{\text{rec}}
    =
    \frac{1}{TD}\sum_{t=1}^{T}\sum_{d=1}^{D}\bigl(X_{t,d}-\hat{X}_{t,d}\bigr)^2,
    \label{eq:rec_loss}
\end{equation}
 This term encourages the temporal latents $\{z_{t,\tau}\}$ and the correlation latent $z_c$ together, via $g_{\theta}$, to preserve the observed multivariate temporal patterns under normal conditions.
 
\noindent
\textbf{Correlation Reconstruction Loss.} 
In multivariate windows, normality is characterized not only by individual pairwise channel relations, but also by the global dependency pattern they form together. If a reconstruction objective matches only entry-wise correlation values, it may miss coherent shifts in the overall dependency structure, whereas a purely structural criterion may overlook localized pairwise breaks. Hence we supervise the correlation latent $z_c$ through the predicted  dependency vector $\hat{r}(X)=h_{\theta}(z_c)$  with a composite loss that targets both aspects. The target $r(X) \in \mathbb{R}^{M}$ carries the lower-triangular entries of the within-window Pearson correlation matrix, where $ M=D(D-1)/2,$ we define
\begin{equation}
    \mathcal{L}_{\text{corr}}
    =
    \mathcal{L}_{\text{Huber}}\!\left(\hat{r}(X), r(X)\right)
    +
    \beta\bigl(1 - \mathrm{cos}(\hat{r}(X), r(X))\bigr),
    \label{eq:corr_loss}
\end{equation}
where $\beta = 0.1$. The Huber term provides entry-wise matching of pairwise correlations,
 \begin{equation}
    \mathcal{L}_{\text{Huber}}(a,b)
    =
    \frac{1}{M}\sum_{m=1}^{M}
    \begin{cases}
        \tfrac{1}{2}(a_m-b_m)^2, & |a_m-b_m|\le \delta,\\[4pt]
        \delta\bigl(|a_m-b_m|-\tfrac{1}{2}\delta\bigr), & |a_m-b_m|> \delta,
    \end{cases}
    \label{eq:huber}
\end{equation}
where $\delta=0.5$. This ensures that small errors below $\delta$ benefit from precise, smooth quadratic gradients, while large errors above $\delta$ are capped linearly, preventing extreme outliers from dominating the reconstruction objective. The cosine term, 
\begin{equation}
    \mathrm{cos}(\hat{r}(X), r(X))
    =
    \frac{\langle \hat{r}(X), r(X)\rangle}{\lVert\hat{r}(X)\rVert_2\,\lVert r(X)\rVert_2},
\end{equation}
forces $\hat{r}(X)$ to preserve the global correlation signature of ${r}(X)$ even when magnitudes are imperfectly matched.

\noindent
\textbf{KL Regularization.} As in a standard variational autoencoder, we add a KL penalty to keep each latent distribution close to a simple Gaussian prior. Without this term, the model can spread probability mass too freely and memorize training windows instead of learning a compact notion of normality. For a latent $z \in \mathbb{R}^{d_z}$ with posterior $q_{\phi}(z \mid X)=\mathcal{N}(z \mid \mu, \mathrm{diag}(\sigma^2))$ and prior $p(z)=\mathcal{N}(0,\sigma_0^2 I)$, we use the standard closed-form KL divergence
\begin{equation}
    \mathcal{L}_{\text{KL}}(z)
    =
    \frac{1}{2}\sum_{j=1}^{d_z}
    \left[
        \log \sigma_0^2
        - \log \sigma_j^2
        + \frac{\sigma_j^2}{\sigma_0^2}
        + \frac{\mu_j^2}{\sigma_0^2}
        - 1
    \right],
    \label{eq:kl_general}
\end{equation}
where $\mu$ and $\log\sigma_j^2$ are the outputs of the corresponding encoder head in~\eqref{eq:encoder_heads}. In training, the sum over $j$ is averaged over latent dimensions.

We apply~\eqref{eq:kl_general} to both branches with different prior scales. At each timestep $\tau$, the temporal latent $z_{t,\tau}$ is regularized toward $p(z_{t,\tau})=\mathcal{N}(0,\sigma_{t,0}^2 I)$ with $\sigma_{t,0}=0.5$. The per-step KL terms are averaged to form $\mathcal{L}_{\text{KL}}^{(t)}$. This tighter prior encourages compact step-level dynamics. The window-level correlation latent uses $\mathcal{L}_{\text{KL}}^{(c)}=\mathcal{L}_{\text{KL}}(z_c)$ with $p(z_c)=\mathcal{N}(0,\sigma_{c,0}^2 I)$ and $\sigma_{c,0}=1.0$.

\noindent
\textbf{Overall Objective.} CARMA is trained by minimizing the weighted sum of all four terms,
\begin{equation}
    \mathcal{L}_{\text{total}}
    =
    \gamma_{\text{rec}}\mathcal{L}_{\text{rec}}
    + \gamma_{\text{corr}}\mathcal{L}_{\text{corr}}
    + \lambda_t \mathcal{L}_{\text{KL}}^{(t)}
    + \lambda_c \mathcal{L}_{\text{KL}}^{(c)},
    \label{eq:total_loss}
\end{equation}
with $\gamma_{\text{rec}}=\gamma_{\text{corr}}=10$, $\lambda_t=0.5$, and $\lambda_c=0.05$. These values were chosen empirically on the TSB-AD-M tuning split and kept fixed across all evaluation series. 
During the first 20 epochs, $\lambda_t$ and $\lambda_c$ are linearly ramped from zero to these final values. The reconstruction term encourages the temporal latents and $z_c$, through $g_{\theta}$, to recover normal multivariate temporal patterns, while $\mathcal{L}_{\text{corr}}$ forces $z_c$ to encode window-level dependency structure.

\textbf{Scoring.} For each stride-1 window we take the reconstruction error at the last observed time step, averaged over channels, and assign it to that global index. We call the resulting series the \emph{pointwise} score. The first $W-1$ test steps are not covered by any full window ending at that step, so we set their scores to zero. Since strong TSB-AD baselines already report window-level scores whose responses are smoothed by overlap, we also evaluate a smoothed CARMA score in the main benchmark comparison. We calculate \emph{smoothed} CARMA score obtained by applying a centered moving average with a uniform kernel of length $W$, matching the training window size. We apply a length-$W$ boxcar filter to the pointwise track and pad symmetrically so the output stays aligned with the original timeline. Throughout the paper we report this smoothed score. 
 \section{Experimental Setup}
 \label{experimental_setup}
 \subsection{Datasets}
 \noindent
 \textbf{TSB-AD-M (Standard benchmark).} For the main benchmark, we use the multivariate split of TSB-AD (TSB-AD-M) \cite{liu2024Elephant}. TSB-AD-M contains 180 evaluation series extracted from diverse real-world domains, including spacecraft telemetry (MSL, SMAP), server monitoring (SMD), medical signals (MIT-BIH), and industrial systems (SWaT, GHL, PSM). Series differ in length, channel count, and anomaly prevalence, so aggregate performance reflects robustness across regimes rather than fit to a single data source. Unlike earlier collections assembled from raw public releases, TSB-AD applies a systematic curation to remove unreliable labels, inconsistent pre-processing, and ambiguous anomaly segments. We refer readers to \cite{liu2024Elephant} for a complete description of the curation pipeline and benchmark design.

\noindent
 \textbf{Synthetic benchmark.} 
 Standard multivariate benchmarks such as TSB-AD-M mix univariate and cross-channel events. Recent work by \cite{pinet2026Anomalies} introduce a segment-level diagnostic that separates univariate, cross-channel, and mixed anomalies on multivariate series, and they propose \textsc{NoiseFlip} and \textsc{NoiseRoll} as synthetic generators in which anomalies disturb the dependence structure between the channels while per-channel marginals remain essentially unchanged. We complement our analysis with experiments on the \textsc{NoiseFlip} and \textsc{NoiseRoll} synthetic datasets of Pinet et al. 
These datasets are created using phase-shifted sinusoidal signals that have common additive noise added to them in order to produce strong baseline correlations. Anomalies are then injected either by flipping the sign of the shared noise component \textsc{NoiseFlip} or by applying a random circular temporal shift to a subset of channels \textsc{NoiseRoll}. Both of these methods maintain the marginal distribution of each individual channel while at the same time disrupting the dependence structure between the channels. We refer readers to \cite{pinet2026Anomalies} work for the background on the original \textsc{NoiseFlip} and \textsc{NoiseRoll} signal generation and anomaly injection mechanisms, and we provide the parameters and generation details used in our experiments in Appendix~\ref{app:synthetic}.

\noindent
\subsection{Baselines and evaluation metrics.} We compare CARMA against classical detectors, reconstruction and forecasting models, and Transformer based methods implemented in the official TSB-AD framework \cite{liu2024Elephant}. We also include methods (MSCRED \cite{chuxuzhangDeepNeuralNetwork}, MTAD\_GAT~\cite{zhao2020Multivariate}, GDN~\cite{deng2021Graph}, CST\_GL~\cite{zhengCorrelationAwareSpatialTemporal2024}) that model inter-channel dependence explicitly. We report threshold-free ranking with AUC-ROC, AUC-PR, VUS-ROC, and VUS-PR, and we report localization with Standard-F1, PA-F1, Event-based-F1, R-based-F1, and Affiliation-F. Our primary metric for ranking methods is VUS-PR, which extends PR evaluation to range anomalies by allowing temporal tolerance near event boundaries \cite{boniol2025VUS}. Liu and Paparrizos analyze the reliability of common TSAD metrics on TSB-AD and argue that several widely used measures are sensitive to label misalignment \cite{liu2024Elephant}. Following their recommendation for TSB-AD, we use VUS-PR for main comparisons and critical-difference diagrams and report the remaining metrics for completeness. Detailed definitions of all evaluation metrics are provided in Appendix~\ref{app:eval_explain}.
 \noindent

 \noindent
 \subsection{Implementation Details} We implement CARMA in PyTorch and evaluate all baseline methods with the official TSB-AD code base. On TSB-AD-M, we use the train/test split provided in each file name and report results on the 180-series evaluation split. On the synthetic dataset, we use the same CARMA configuration and the same evaluation code. CARMA is trained with Adam for 100 epochs using learning rate $10^{-4}$, batch size 128, and input window size 100. At test time, anomaly scores are computed from reconstruction error only, unless stated otherwise. For every method in our main comparison, we repeat training with seeds 2024, 2025, and 2026.
 
\section{Results}

\subsection{TSB-AD-M Benchmark Results}
Table~\ref{tab:leaderboard} compares CARMA with 29 multivariate baselines, reporting the mean over 180 series and three runs. CARMA achieves a mean VUS-PR of $0.4929$, ahead of the strongest baseline, HYDRA Approx ($0.4409$), by $0.052$ absolute points ($11.8\%$ relative). It is also best on AUC-PR ($0.4425$), AUC-ROC ($0.8330$), VUS-ROC ($0.8598$), Standard-F1 ($0.5301$), and R-based-F1 ($0.5088$). The range-aware metrics (VUS-PR, VUS-ROC, R-based-F1) reward scores that stay elevated across anomalous intervals rather than reacting to brief local fluctuations. On Affiliation-F and Event-based-F1, CARMA is second, $0.0051$ and $0.0237$ below MTAD-GAT respectively. MTAD-GAT scores anomalies from short-horizon forecast errors, which suits event-level metrics that reward detecting at least part of each anomalous event, but it is not competitive on the threshold-free and range-aware metrics where CARMA exceeds it by $0.142$ in VUS-PR (about $41\%$ relative) and by $0.08$ to $0.12$ on every other such metric, and MTAD-GAT ranks only 12th among the baselines on VUS-PR. CARMA does not lead on PA-F1. However, point adjustment credits an entire anomalous segment once any point in it is detected, which is known to inflate results substantially \cite{kimRigorousEvaluationTimeseries2022}, so we report it for completeness only. We have reported the standard deviation over random seeds and the spread over the time series in Appendix \ref{app:tsbadm_analysis}.

\begin{figure}
    \centering
    \begin{minipage}[c]{0.70\linewidth}
        \centering
        \includegraphics[width=\linewidth]{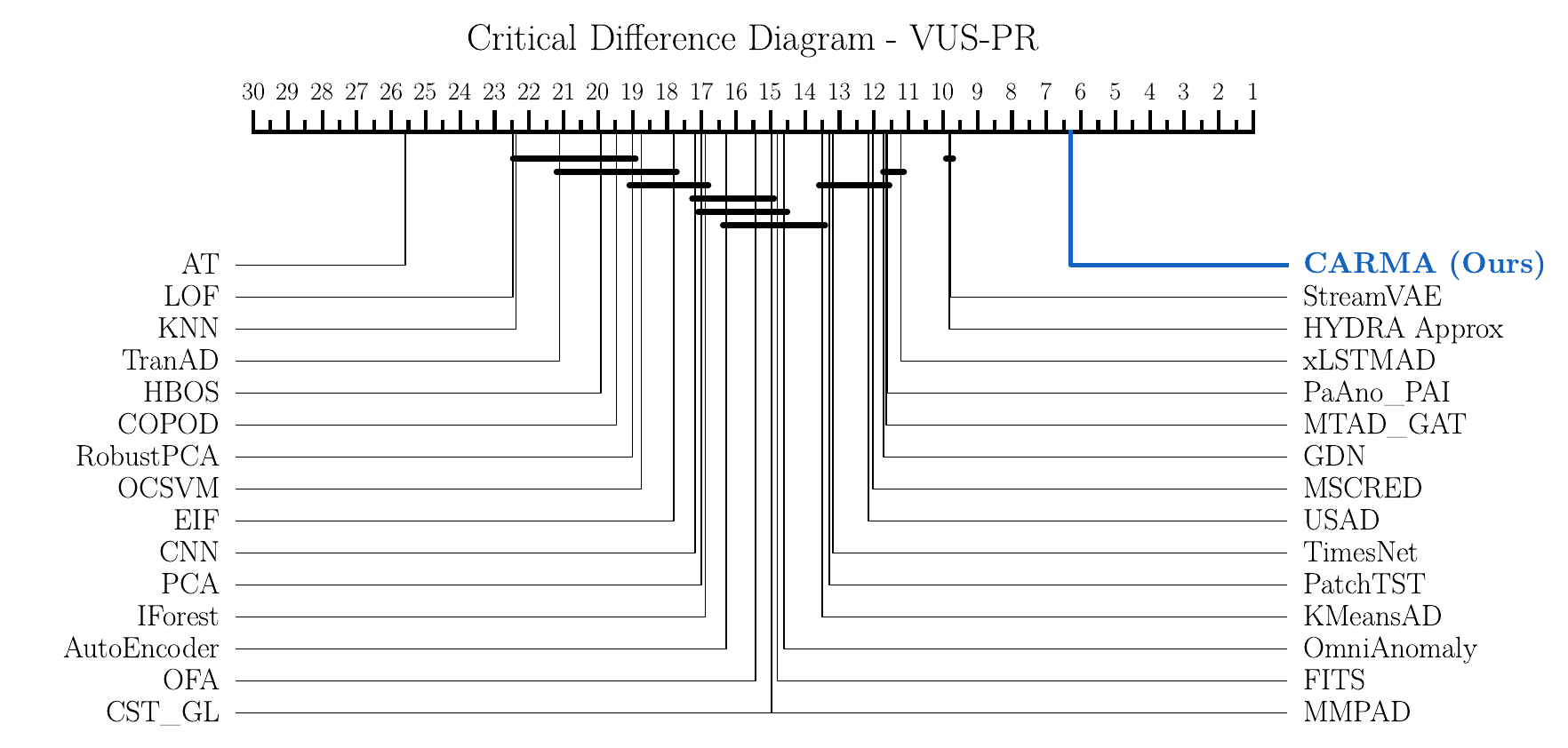}
    \end{minipage}%
    \hfill
    \begin{minipage}[c]{0.29\linewidth}
        \centering
        \includegraphics[width=\linewidth]{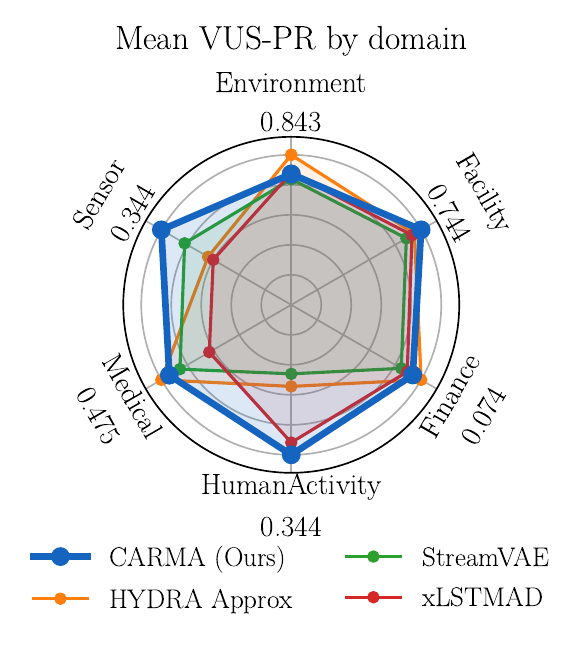}
    \end{minipage}
    \caption{Critical difference diagrams on TSB-AD-M for VUS-PR (left) and domain-specific performance (right). Models connected by a horizontal bar are not significantly different (Wilcoxon signed-rank test, 0.05). Lower average rank indicates better performance.}
    \label{fig:cd_tsbad_combined}
\end{figure}

\textbf{Statistical Comparison.} Figure~\ref{fig:cd_tsbad_combined} shows the VUS-PR critical-difference diagram over 180 series (left) and mean VUS-PR by application domain for the leading methods (right). 
In this diagram, models are not statistically significant if they are in the same clique (the black bar). For the statistical test, we used the Wilcoxon signed-rank test with Holm correction as the post hoc test to the Friedman test~\cite{demsarStatisticalComparisonsClassifiers2006}. CARMA has the best average rank and is not connected to any other method so its VUS-PR advantage is statistically significant over the full 180-series comparison. The strongest alternatives, including HYDRA Approx and StreamVAE, are statistically tied among themselves. The domain panel shows CARMA performs well on the Sensor, Human Activity and Facility domain, and remains competitive elsewhere without collapsing to a single-domain effect.

\begin{table}
\centering
\caption{Results on the TSB-AD-M benchmark. Mean of each metric over 180 series and three seeds. The best and second-best results in each column are given in bold and underlined, respectivel. Where AT and AE refer to Anomaly Transformer and AutoEncoder, respectively.}
\label{tab:leaderboard}
\fontsize{7pt}{9pt}\selectfont  
\setlength{\tabcolsep}{6pt}
\renewcommand{\arraystretch}{1}  
\begin{tabular}{>{\centering\arraybackslash}p{0.24\linewidth} 
                >{\centering\arraybackslash}p{0.04\linewidth} 
                >{\centering\arraybackslash}p{0.04\linewidth} 
                >{\centering\arraybackslash}p{0.04\linewidth} 
                >{\centering\arraybackslash}p{0.04\linewidth} 
                >{\centering\arraybackslash}p{0.05\linewidth}
                >{\centering\arraybackslash}p{0.04\linewidth} 
                >{\centering\arraybackslash}p{0.07\linewidth} 
                >{\centering\arraybackslash}p{0.07\linewidth}
                >{\centering\arraybackslash}p{0.07\linewidth}
                }
\toprule
Method & AUC-PR & AUC-ROC & VUS-PR & VUS-ROC & Standard-F1 & PA-F1 & Event-based-F1 & R-based-F1 & Affiliation-F \\
\midrule
HYDRA~\cite{huang2026HYDRA} & 0.3933 & 0.7727 & \underline{0.4409} & \underline{0.8068} & 0.4509 & 0.6783 & 0.5669 & \underline{0.4303} & 0.8556 \\
StreamVAE~\cite{ozer2026STREAMVAE} & 0.3881 & 0.7578 & 0.4277 & 0.7740 & \underline{0.4551} & 0.5390 & 0.4757 & 0.4162 & 0.8311 \\
PaAno\_PAI~\cite{parkPaAnoPatchBasedRepresentation2026} & \underline{0.3992} & \underline{0.7834} & 0.4049 & 0.8059 & 0.4345 & 0.7732 & 0.6109 & 0.3804 & 0.8679 \\
USAD~\cite{audibertUSADUnSupervisedAnomaly2020} & 0.3625 & 0.7366 & 0.4026 & 0.7652 & 0.4329 & 0.6229 & 0.5189 & 0.4227 & 0.8366 \\
TimesNet~\cite{wuTimesNetTemporal2DVariation2023} & 0.3375 & 0.6697 & 0.3910 & 0.7103 & 0.4115 & 0.6691 & 0.5336 & 0.4076 & 0.8476 \\
PatchTST~\cite{nieTimeSeriesWorth2023} & 0.3360 & 0.6746 & 0.3902 & 0.7167 & 0.4067 & 0.7097 & 0.5355 & 0.3777 & 0.8460 \\
xLSTMAD~\cite{faberXLSTMADPowerfulXLSTMbased2025} & 0.3420 & 0.7415 & 0.3895 & 0.7804 & 0.4254 & 0.6195 & 0.5330 & 0.3910 & 0.8325 \\
FITS~\cite{xuFITSModelingTime2024} & 0.3053 & 0.6509 & 0.3642 & 0.6994 & 0.3702 & 0.7359 & 0.5011 & 0.3220 & 0.8429 \\
MSCRED~\cite{chuxuzhangDeepNeuralNetwork} & 0.3500 & 0.7123 & 0.3622 & 0.7365 & 0.3989 & 0.5987 & 0.5054 & 0.4071 & 0.8236 \\
OFA~\cite{zhouOneFitsAll2023} & 0.3005 & 0.6397 & 0.3557 & 0.6887 & 0.3692 & 0.6727 & 0.5188 & 0.3620 & 0.8350 \\
MMPAD~\cite{yehMatrixProfileTimeSeries2026} & 0.3091 & 0.7444 & 0.3532 & 0.7582 & 0.3581 & 0.4786 & 0.3901 & 0.2997 & 0.8129 \\
MTAD\_GAT~\cite{zhao2020Multivariate} & 0.3648 & 0.7361 & 0.3506 & 0.7608 & 0.4143 & \underline{0.7866} & \textbf{0.6454} & 0.3836 & \textbf{0.8797} \\
GDN~\cite{deng2021Graph} & 0.3106 & 0.7253 & 0.3318 & 0.7578 & 0.3684 & 0.7265 & 0.5673 & 0.3728 & 0.8602 \\
OmniAnomaly~\cite{suRobustAnomalyDetection2019} & 0.3152 & 0.7058 & 0.3153 & 0.7435 & 0.3752 & \textbf{0.7870} & 0.6047 & 0.2936 & 0.8490 \\
CNN~\cite{munirDeepAnTDeepLearning2019} & 0.2934 & 0.6446 & 0.2960 & 0.6728 & 0.3461 & 0.6884 & 0.5376 & 0.2516 & 0.7215 \\
CST\_GL~\cite{zhengCorrelationAwareSpatialTemporal2024} & 0.2771 & 0.6787 & 0.2937 & 0.7120 & 0.3321 & 0.7160 & 0.5331 & 0.3358 & 0.8429 \\
KMeansAD~\cite{aggarwalOutlierAnalysis2017} & 0.2485 & 0.6869 & 0.2867 & 0.7241 & 0.3052 & 0.6723 & 0.4840 & 0.3269 & 0.8209 \\
AE~\cite{kingma2022AutoEncoding} & 0.2635 & 0.6674 & 0.2835 & 0.6951 & 0.3208 & 0.5924 & 0.4322 & 0.2856 & 0.8185 \\
PCA~\cite{aggarwalOutlierAnalysis2017} & 0.2409 & 0.6759 & 0.2691 & 0.7058 & 0.3120 & 0.5118 & 0.3698 & 0.3277 & 0.7887 \\
TranAD~\cite{tuli2022TranAD} & 0.2654 & 0.5922 & 0.2651 & 0.6214 & 0.3192 & 0.6569 & 0.5053 & 0.2360 & 0.7058 \\
OCSVM~\cite{scholkopfEstimatingSupportHighdimensional2001} & 0.2354 & 0.6341 & 0.2628 & 0.6697 & 0.2890 & 0.6351 & 0.4416 & 0.3142 & 0.8169 \\
IForest~\cite{liuIsolationForest2008} & 0.2095 & 0.7040 & 0.2487 & 0.7437 & 0.2798 & 0.6553 & 0.4027 & 0.2434 & 0.8010 \\
RobustPCA~\cite{xuRobustPCAOutlier2010} & 0.2386 & 0.5898 & 0.2379 & 0.6122 & 0.2833 & 0.5739 & 0.3788 & 0.3301 & 0.7919 \\
EIF~\cite{haririExtendedIsolationForest2021} & 0.1850 & 0.6685 & 0.2052 & 0.7057 & 0.2574 & 0.7388 & 0.4298 & 0.2561 & 0.8112 \\
COPOD~\cite{liCOPODCopulaBasedOutlier2020} & 0.2046 & 0.6518 & 0.2015 & 0.6820 & 0.2688 & 0.7170 & 0.4137 & 0.2422 & 0.7991 \\
HBOS~\cite{dengelHistogrambasedOutlierScore} & 0.1610 & 0.6331 & 0.1856 & 0.6691 & 0.2439 & 0.6670 & 0.3990 & 0.2438 & 0.7963 \\
KNN~\cite{ramaswamyEfficientAlgorithmsMining2000} & 0.1317 & 0.4961 & 0.1693 & 0.5667 & 0.1826 & 0.6764 & 0.4410 & 0.2048 & 0.7895 \\
LOF~\cite{breunigLOFIdentifyingDensitybased2000} & 0.0953 & 0.5321 & 0.1310 & 0.5831 & 0.1493 & 0.5600 & 0.3250 & 0.1474 & 0.7635 \\
AT~\cite{xu2022Anomaly} & 0.0643 & 0.3978 & 0.0989 & 0.4286 & 0.1021 & 0.5186 & 0.3410 & 0.1319 & 0.6353 \\
CARMA (Ours) & \textbf{0.4425} & \textbf{0.8330} & \textbf{0.4929} & \textbf{0.8598} & \textbf{0.5301} & 0.7255 & \underline{0.6217} & \textbf{0.5088} & \underline{0.8746} \\
\bottomrule
\end{tabular}
\end{table}

\subsection{Synthetic Data Analysis}
Table~\ref{tab:synthetic_leaderboard} summarizes the results on the synthetic datasets, where anomalies disrupt cross-channel dependence while preserving per-channel marginals. We compare CARMA with the five strongest baselines on TSB-AD-M by mean VUS-PR (HYDRA Approx, StreamVAE, PaAno\_PAI, USAD, and TimesNet), since the synthetic benchmark is intended to test whether methods that perform well on real data also capture cross-channel anomalies. CARMA achieves the strongest overall performance, leading in most ranking and event-based metrics and securing the highest VUS-PR of 0.9888. StreamVAE stands out as the closest competitor on AUC and VUS based measures and it models the multivariate series through a single temporal generative pathway, but it does not model how channels relate inside each window. PaAno\_PAI achieves the highest PA-F1, but it severely underperforms on other metrics. 

\begin{table}
\centering
\caption{Comparison with the main benchmark's five strongest baselines on synthetic datasets.}
\label{tab:synthetic_leaderboard}
\fontsize{7pt}{9pt}\selectfont  
\setlength{\tabcolsep}{6pt}
\renewcommand{\arraystretch}{1}  
\begin{tabular}{>{\centering\arraybackslash}p{0.24\linewidth} 
                >{\centering\arraybackslash}p{0.04\linewidth} 
                >{\centering\arraybackslash}p{0.04\linewidth} 
                >{\centering\arraybackslash}p{0.04\linewidth} 
                >{\centering\arraybackslash}p{0.04\linewidth} 
                >{\centering\arraybackslash}p{0.05\linewidth}
                >{\centering\arraybackslash}p{0.04\linewidth} 
                >{\centering\arraybackslash}p{0.07\linewidth} 
                >{\centering\arraybackslash}p{0.07\linewidth}
                >{\centering\arraybackslash}p{0.07\linewidth}
                }
\toprule
Method & AUC-PR & AUC-ROC & VUS-PR & VUS-ROC & Standard-F1 & PA-F1 & Event-based-F1 & R-based-F1 & Affiliation-F \\
\midrule
StreamVAE~\cite{ozer2026STREAMVAE} & \underline{0.9716} & \underline{0.9962} & \underline{0.9858} & \underline{0.9982} & \underline{0.9245} & 0.9389 & \underline{0.9375} & \underline{0.9213} & \underline{0.9838} \\
HYDRA~\cite{huang2026HYDRA} & 0.8681 & 0.9382 & 0.8681 & 0.9500 & 0.8180 & 0.9200 & 0.9183 & 0.7785 & 0.9376 \\
PaAno\_PAI~\cite{parkPaAnoPatchBasedRepresentation2026} & 0.7336 & 0.9300 & 0.7559 & 0.9451 & 0.6797 & \textbf{0.9807} & 0.9361 & 0.3342 & 0.9358 \\
TimesNet~\cite{wuTimesNetTemporal2DVariation2023} & 0.5300 & 0.7588 & 0.5510 & 0.7849 & 0.5417 & 0.8408 & 0.6504 & 0.5473 & 0.8432 \\
USAD~\cite{audibertUSADUnSupervisedAnomaly2020} & 0.5104 & 0.7402 & 0.5322 & 0.7638 & 0.5287 & 0.8785 & 0.6003 & 0.4479 & 0.8337 \\
CARMA (Ours) & \textbf{0.9768} & \textbf{0.9968} & \textbf{0.9888} & \textbf{0.9986} & \textbf{0.9259} & \underline{0.9426} & \textbf{0.9411} & \textbf{0.9329} & \textbf{0.9859} \\

\bottomrule
\end{tabular}
\end{table}

\subsection{Ablation}
\label{sec:abl}

\paragraph{Ablation Variants.} To separate the effect of adding a second latent from the effect of supervising it, we compare three models that differ only in the correlation branch. In Table~\ref{tab:comp_abl}, the \emph{Single latent} (w/o $z_c$) variant removes the correlation branch entirely by conditioning the series decoder on $z_t$ alone and omitting both $\mathcal{L}_{\text{corr}}$ and $\mathcal{L}_{\text{KL}}^{(c)}$ terms. \emph{Unsupervised $z_c$} (w/o $\mathcal{L}_{\text{corr}}$) keeps the window-level latent, its KL term, and its injection into the series decoder, but removes $\mathcal{L}_{\text{corr}}$, so $z_c$ is free to encode whatever helps reconstruction. This mirrors unsupervised dual-latent designs such as InterFusion. The remaining rows of Table~\ref{tab:comp_abl} change one setting at a time relative to CARMA: the correlation target (Spearman instead of Pearson; Pearson at lags $-5$ to $+5$ instead of lag 0) and the temporal prior ($\sigma_{t,0}=1$ instead of $0.5$). 

\paragraph{Component Contribution.} Table~\ref{tab:comp_abl} reports mean and standard deviation over three seeds on TSB-AD-M. Moving from the single-latent model to unsupervised $z_c$ increases VUS-PR from 0.4620 to 0.4719 and adding supervision raises it further to 0.4929 with this ordering holding across every metric. Most of the improvement therefore comes from what $z_c$ is trained to encode, not from the extra latent itself. The lag-5 variant falls below even the single-latent model (VUS-PR $0.4042$). Pearson is more effective than Spearman correlation. Setting the temporal prior to $\sigma_{t,0}=1$ is close to CARMA ($0.4881$) but consistently below it, and CARMA also has the smallest seed variance. In Appendix~\ref{app:abl_carma_vs_nocorr}, we present a case study comparing CARMA against the single-latent variant (w/o~$z_c$) and visualize their learned embedding spaces.

\begin{table}
\centering
\caption{CARMA ablation study reported as mean $\pm$ standard deviation over seeds. The best and second-best results in each column are given in bold and underlined, respectively.} 
\label{tab:comp_abl}
\fontsize{6.5pt}{9pt}\selectfont  
\setlength{\tabcolsep}{6pt}
\renewcommand{\arraystretch}{1.1}  
\begin{tabular}{>{\centering\arraybackslash}p{0.20\linewidth} 
                >{\centering\arraybackslash}p{0.12\linewidth} 
                >{\centering\arraybackslash}p{0.12\linewidth} 
                >{\centering\arraybackslash}p{0.12\linewidth} 
                >{\centering\arraybackslash}p{0.12\linewidth} 
                >{\centering\arraybackslash}p{0.12\linewidth}
                }
\toprule
Method & AUC-PR & AUC-ROC & VUS-PR & VUS-ROC & Standard-F1 \\ 
\midrule
CARMA + Pearson lag=5 & 0.3468 $\pm$ 0.0069 & 0.7985 $\pm$ 0.0023 & 0.4042 $\pm$ 0.0057 & 0.8376 $\pm$ 0.0018 & 0.4315 $\pm$ 0.0067 \\
CARMA w/o $z_c$ & 0.4180 $\pm$ 0.0047 & 0.8041 $\pm$ 0.0034 & 0.4620 $\pm$ 0.0032 & 0.8287 $\pm$ 0.0036 & 0.4936 $\pm$ 0.0053 \\
CARMA w/o $\mathcal{L}_{\text{corr}}$ & 0.4243 $\pm$ 0.0060 & 0.8157 $\pm$ 0.0032 & 0.4719 $\pm$ 0.0055 & 0.8424 $\pm$ 0.0031 & 0.5033 $\pm$ 0.0084 \\
CARMA + Spearman corr & 0.4302 $\pm$ 0.0033 & 0.8218 $\pm$ 0.0007 & 0.4780 $\pm$ 0.0035 & 0.8488 $\pm$ 0.0006 & 0.5102 $\pm$ 0.0030 \\
CARMA $\sigma_{t,0}=1$ & \underline{0.4384} $\pm$ 0.0021 & \underline{0.8313} $\pm$ 0.0059 & \underline{0.4881} $\pm$ 0.0033 & \underline{0.8575} $\pm$ 0.0066 & \underline{0.5271} $\pm$ 0.0038 \\
CARMA (Ours) & \textbf{0.4425} $\pm$ 0.0009 & \textbf{0.8330} $\pm$ 0.0017 & \textbf{0.4929} $\pm$ 0.0027 & \textbf{0.8598} $\pm$ 0.0010 & \textbf{0.5301} $\pm$ 0.0032 \\
\bottomrule
\end{tabular}
\end{table}

\begin{figure}
    \centering
    \includegraphics[width=0.8\linewidth]{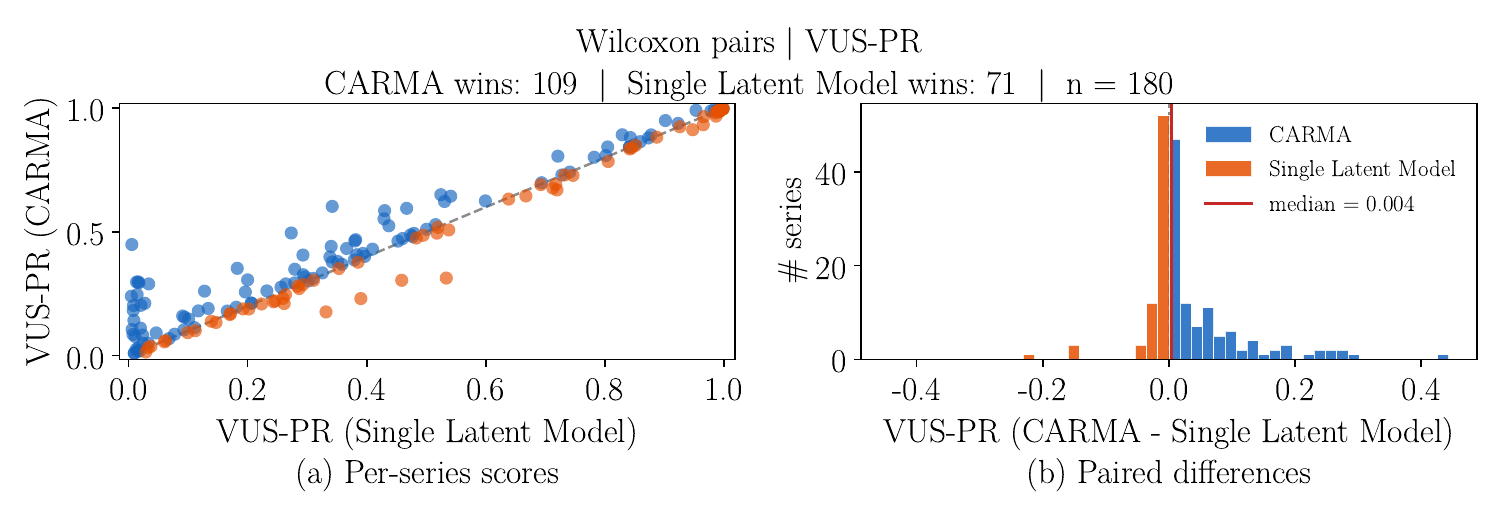}
\caption{Paired pointwise VUS-PR on TSB-AD-M for CARMA and single latent model. Panel (a) compares scores for all 180 series, with points above the diagonal favoring the correlated model. Panel (b) shows the paired differences.}
    \label{fig:wilcoxon-correlation}
\end{figure}

\textbf{Comparison with the single-latent model.}
\label{sec:w/o ablation}
We performed a Wilcoxon signed-rank test to compare CARMA with the single latent model on TSB-AD-M. The paired scores tend to favor the correlated model, which wins on 109 of the 180 series (Wilcoxon signed-rank, $p = 9.7 \times 10^{-7}$). As shown in Figure~\ref{fig:wilcoxon-correlation} (b), the differences are small and cluster around zero, with a tail extending toward the correlated model. The median difference is $0.004$ in favor of the correlated model, so a typical series performs nearly the same under both variants while a smaller subset shows a larger gain.

\paragraph{Reconstruction error at anomalies.} We compare CARMA with the single-latent model on pointwise reconstruction error averaged over the anomalous timesteps of each series and tested with a paired Wilcoxon signed-rank test across the 180 series. CARMA produces higher error at anomalies on 132 of 180 series ($p = 3.2 \times 10^{-10}$), consistent with a stronger error response for anomalies.
\section{Conclusion}

Reconstruction-based multivariate anomaly detectors typically learn normal behavior through shared encoders that mix channels implicitly. They optimize a single signal reconstruction objective, so inter-channel correlation is never modeled as an explicit target. As a result, flexible decoders can reconstruct plausible per-channel trajectories even when cross-channel correlations are abnormal, and correlation-breaking events can produce weak anomaly scores. CARMA addresses this by keeping standard series reconstruction and adding a second reconstruction problem in the space of inter-channel correlations. On TSB-AD-M, CARMA achieves the best mean VUS-PR (0.4929) among 29 baselines, and removing the correlation reconstruction module degrades performance on every reported metric. A paired ablation over all evaluation series confirms that the gain is systematic rather than dataset-specific ($p = 9.7 \times 10^{-7}$). On synthetic NoiseFlip and NoiseRoll benchmarks, where anomalies disrupt correlation while preserving marginal channel behavior, CARMA remains competitive with strong baseline methods.


\newpage
\bibliographystyle{plain}
\bibliography{MyLibrarybibtex}

\newpage
\section*{Appendix}

\subsection{Analysis on TSB-AD-M Benchmark}
\label{app:tsbadm_analysis}
\textbf{Spread Over Series.} Table~\ref{tab:std_spread} shows that detection quality varies substantially across the 180 TSB-AD-M series. For nearly every method the spread is large relative to the typical score which means strong results on some files coexist with weak results on others. Both deep and classical detectors show this pattern. Among competitive baselines, CARMA’s cross-series variability is comparable to that of HYDRA Approx, StreamVAE, and PaAno\_PAI. Even against strong baselines like HYDRA Approx and StreamVAE, CARMA maintains a relatively lower spread ($\pm$ 0.3287) while achieving the highest mean VUS-PR (0.4929).

\textbf{Spread Over Seeds.} Table~\ref{tab:std_seed} reports the variation in macro-averaged scores across training seeds 2024, 2025, and 2026. Some detectors shows small but measurable seed effects where StreamVAE and PaAno\_PAI remain stable among strong baselines while CNN and TranAD fluctuate more noticeably between seeds. CARMA achieves the highest VUS-PR of $0.4929$ with a standard deviation of $0.0027$ across three seeds.

\begin{table}[H]
\centering
\caption{Results on the TSB-AD-M benchmark. Mean $\pm$ standard deviation over all evaluated series (spread). The best and second-best results in each column are given in bold and underlined, respectively. Where AT and AE refer to Anomaly Transformer and AutoEncoder, respectively.} 
\label{tab:std_spread}
\fontsize{6.5pt}{9pt}\selectfont  
\setlength{\tabcolsep}{6pt}
\renewcommand{\arraystretch}{1.1}  
\begin{tabular}{>{\centering\arraybackslash}p{0.22\linewidth} 
                >{\centering\arraybackslash}p{0.12\linewidth} 
                >{\centering\arraybackslash}p{0.12\linewidth} 
                >{\centering\arraybackslash}p{0.12\linewidth} 
                >{\centering\arraybackslash}p{0.12\linewidth} 
                >{\centering\arraybackslash}p{0.12\linewidth}
                }
\toprule
Method & AUC-PR & AUC-ROC & VUS-PR & VUS-ROC & Standard-F1 \\ 
\midrule
HYDRA~\cite{huang2026HYDRA} & 0.3933 $\pm$ 0.3208 & 0.7727 $\pm$ 0.2245 & \underline{0.4409} $\pm$ 0.3409 & \underline{0.8068} $\pm$ 0.2144 & 0.4509 $\pm$ 0.3112 \\
StreamVAE~\cite{ozer2026STREAMVAE} & 0.3881 $\pm$ 0.3539 & 0.7578 $\pm$ 0.2382 & 0.4277 $\pm$ 0.3635 & 0.7740 $\pm$ 0.2363 & \underline{0.4551} $\pm$ 0.3367 \\
PaAno\_PAI~\cite{parkPaAnoPatchBasedRepresentation2026} & \underline{0.3992} $\pm$ 0.3370 & \underline{0.7834} $\pm$ 0.2133 & 0.4049 $\pm$ 0.3520 & 0.8059 $\pm$ 0.2070 & 0.4345 $\pm$ 0.3130 \\
USAD~\cite{audibertUSADUnSupervisedAnomaly2020} & 0.3625 $\pm$ 0.3392 & 0.7366 $\pm$ 0.2562 & 0.4026 $\pm$ 0.3490 & 0.7652 $\pm$ 0.2459 & 0.4329 $\pm$ 0.3227 \\
TimesNet~\cite{wuTimesNetTemporal2DVariation2023} & 0.3375 $\pm$ 0.3119 & 0.6697 $\pm$ 0.3028 & 0.3910 $\pm$ 0.3373 & 0.7103 $\pm$ 0.2968 & 0.4115 $\pm$ 0.3128 \\
PatchTST~\cite{nieTimeSeriesWorth2023} & 0.3360 $\pm$ 0.3101 & 0.6746 $\pm$ 0.3049 & 0.3902 $\pm$ 0.3357 & 0.7167 $\pm$ 0.2992 & 0.4067 $\pm$ 0.3119 \\
xLSTMAD~\cite{faberXLSTMADPowerfulXLSTMbased2025} & 0.3420 $\pm$ 0.3152 & 0.7415 $\pm$ 0.2228 & 0.3895 $\pm$ 0.3336 & 0.7804 $\pm$ 0.2177 & 0.4254 $\pm$ 0.2984 \\
FITS~\cite{xuFITSModelingTime2024} & 0.3053 $\pm$ 0.3232 & 0.6509 $\pm$ 0.3043 & 0.3642 $\pm$ 0.3518 & 0.6994 $\pm$ 0.2995 & 0.3702 $\pm$ 0.3197 \\
MSCRED~\cite{chuxuzhangDeepNeuralNetwork} & 0.3500 $\pm$ 0.3610 & 0.7123 $\pm$ 0.2188 & 0.3622 $\pm$ 0.3721 & 0.7365 $\pm$ 0.2206 & 0.3989 $\pm$ 0.3445 \\
OFA~\cite{zhouOneFitsAll2023} & 0.3005 $\pm$ 0.3013 & 0.6397 $\pm$ 0.2877 & 0.3557 $\pm$ 0.3368 & 0.6887 $\pm$ 0.2853 & 0.3692 $\pm$ 0.3027 \\
MMPAD~\cite{yehMatrixProfileTimeSeries2026} & 0.3091 $\pm$ 0.3323 & 0.7444 $\pm$ 0.2176 & 0.3532 $\pm$ 0.3583 & 0.7582 $\pm$ 0.2174 & 0.3581 $\pm$ 0.3160 \\
MTAD\_GAT~\cite{zhao2020Multivariate} & 0.3648 $\pm$ 0.3737 & 0.7361 $\pm$ 0.2065 & 0.3506 $\pm$ 0.3759 & 0.7608 $\pm$ 0.2032 & 0.4143 $\pm$ 0.3518 \\
GDN~\cite{deng2021Graph} & 0.3106 $\pm$ 0.3188 & 0.7253 $\pm$ 0.1798 & 0.3318 $\pm$ 0.3471 & 0.7578 $\pm$ 0.1793 & 0.3684 $\pm$ 0.3034 \\
OmniAnomaly~\cite{suRobustAnomalyDetection2019} & 0.3152 $\pm$ 0.3643 & 0.7058 $\pm$ 0.2231 & 0.3153 $\pm$ 0.3643 & 0.7435 $\pm$ 0.2098 & 0.3752 $\pm$ 0.3478 \\
CST\_GL~\cite{zhengCorrelationAwareSpatialTemporal2024} & 0.2771 $\pm$ 0.3249 & 0.6787 $\pm$ 0.2140 & 0.2937 $\pm$ 0.3457 & 0.7120 $\pm$ 0.2171 & 0.3321 $\pm$ 0.3106 \\
KMeansAD~\cite{aggarwalOutlierAnalysis2017} & 0.2485 $\pm$ 0.2622 & 0.6869 $\pm$ 0.1976 & 0.2867 $\pm$ 0.2961 & 0.7241 $\pm$ 0.1936 & 0.3052 $\pm$ 0.2450 \\
AE~\cite{kingma2022AutoEncoding} & 0.2635 $\pm$ 0.3314 & 0.6674 $\pm$ 0.2089 & 0.2835 $\pm$ 0.3492 & 0.6951 $\pm$ 0.2086 & 0.3208 $\pm$ 0.3257 \\
PCA~\cite{aggarwalOutlierAnalysis2017} & 0.2409 $\pm$ 0.2909 & 0.6759 $\pm$ 0.2377 & 0.2691 $\pm$ 0.3034 & 0.7058 $\pm$ 0.2249 & 0.3120 $\pm$ 0.2875 \\
CNN~\cite{munirDeepAnTDeepLearning2019} & 0.2641 $\pm$ 0.2658 & 0.5803 $\pm$ 0.1294 & 0.2663 $\pm$ 0.2676 & 0.6056 $\pm$ 0.1201 & 0.3114 $\pm$ 0.2479 \\
OCSVM~\cite{scholkopfEstimatingSupportHighdimensional2001} & 0.2354 $\pm$ 0.3030 & 0.6341 $\pm$ 0.2078 & 0.2628 $\pm$ 0.3229 & 0.6697 $\pm$ 0.2166 & 0.2890 $\pm$ 0.2954 \\
IForest~\cite{liuIsolationForest2008} & 0.2095 $\pm$ 0.2323 & 0.7040 $\pm$ 0.1909 & 0.2487 $\pm$ 0.2601 & 0.7437 $\pm$ 0.1851 & 0.2798 $\pm$ 0.2360 \\
TranAD~\cite{tuli2022TranAD} & 0.2388 $\pm$ 0.2708 & 0.5330 $\pm$ 0.1701 & 0.2386 $\pm$ 0.2710 & 0.5593 $\pm$ 0.1598 & 0.2873 $\pm$ 0.2589 \\
RobustPCA~\cite{xuRobustPCAOutlier2010} & 0.2386 $\pm$ 0.3492 & 0.5898 $\pm$ 0.2407 & 0.2379 $\pm$ 0.3504 & 0.6122 $\pm$ 0.2349 & 0.2833 $\pm$ 0.3340 \\
EIF~\cite{haririExtendedIsolationForest2021} & 0.1850 $\pm$ 0.2176 & 0.6685 $\pm$ 0.1722 & 0.2052 $\pm$ 0.2534 & 0.7057 $\pm$ 0.1670 & 0.2574 $\pm$ 0.2393 \\
COPOD~\cite{liCOPODCopulaBasedOutlier2020} & 0.2046 $\pm$ 0.2919 & 0.6518 $\pm$ 0.1848 & 0.2015 $\pm$ 0.2886 & 0.6820 $\pm$ 0.1777 & 0.2688 $\pm$ 0.2916 \\
HBOS~\cite{dengelHistogrambasedOutlierScore} & 0.1610 $\pm$ 0.1955 & 0.6331 $\pm$ 0.1834 & 0.1856 $\pm$ 0.2418 & 0.6691 $\pm$ 0.1829 & 0.2439 $\pm$ 0.2319 \\
KNN~\cite{ramaswamyEfficientAlgorithmsMining2000} & 0.1317 $\pm$ 0.1553 & 0.4961 $\pm$ 0.2182 & 0.1693 $\pm$ 0.2136 & 0.5667 $\pm$ 0.2154 & 0.1826 $\pm$ 0.1669 \\
LOF~\cite{breunigLOFIdentifyingDensitybased2000} & 0.0953 $\pm$ 0.0920 & 0.5321 $\pm$ 0.0981 & 0.1310 $\pm$ 0.1835 & 0.5831 $\pm$ 0.1415 & 0.1493 $\pm$ 0.1180 \\
AT~\cite{xu2022Anomaly} & 0.0578 $\pm$ 0.0498 & 0.3586 $\pm$ 0.0596 & 0.0888 $\pm$ 0.1353 & 0.3860 $\pm$ 0.0889 & 0.0918 $\pm$ 0.0633 \\
CARMA (Ours) & \textbf{0.4425} $\pm$ 0.3230 & \textbf{0.8330} $\pm$ 0.1778 & \textbf{0.4929} $\pm$ 0.3287 & \textbf{0.8598} $\pm$ 0.1595 & \textbf{0.5301} $\pm$ 0.2868 \\
\bottomrule
\end{tabular}
\end{table}

\begin{table}[t]
\centering
\caption{Results on the TSB-AD-M benchmark. Mean $\pm$ standard deviation over three seeds. The best and second-best results in each column are given in bold and underlined, respectively. Where AT and AE refer to Anomaly Transformer and AutoEncoder, respectively.} 
\label{tab:std_seed}
\fontsize{6.5pt}{9pt}\selectfont  
\setlength{\tabcolsep}{6pt}
\renewcommand{\arraystretch}{1.1}  
\begin{tabular}{>{\centering\arraybackslash}p{0.22\linewidth} 
                >{\centering\arraybackslash}p{0.12\linewidth} 
                >{\centering\arraybackslash}p{0.12\linewidth} 
                >{\centering\arraybackslash}p{0.12\linewidth} 
                >{\centering\arraybackslash}p{0.12\linewidth} 
                >{\centering\arraybackslash}p{0.12\linewidth}
                }
\toprule
Method & AUC-PR & AUC-ROC & VUS-PR & VUS-ROC & Standard-F1 \\ 
\midrule
HYDRA~\cite{huang2026HYDRA} & 0.3933 $\pm$ 0.0056 & 0.7727 $\pm$ 0.0029 & \underline{0.4409} $\pm$ 0.0052 & \underline{0.8068} $\pm$ 0.0028 & 0.4509 $\pm$ 0.0014 \\
StreamVAE~\cite{ozer2026STREAMVAE} & 0.3881 $\pm$ 0.0021 & 0.7578 $\pm$ 0.0021 & 0.4277 $\pm$ 0.0023 & 0.7740 $\pm$ 0.0017 & \underline{0.4551} $\pm$ 0.0006 \\
PaAno\_PAI~\cite{parkPaAnoPatchBasedRepresentation2026} & \underline{0.3992} $\pm$ 0.0000 & \underline{0.7834} $\pm$ 0.0002 & 0.4049 $\pm$ 0.0001 & 0.8059 $\pm$ 0.0002 & 0.4345 $\pm$ 0.0000 \\
USAD~\cite{audibertUSADUnSupervisedAnomaly2020} & 0.3625 $\pm$ 0.0004 & 0.7366 $\pm$ 0.0016 & 0.4026 $\pm$ 0.0008 & 0.7652 $\pm$ 0.0021 & 0.4329 $\pm$ 0.0005 \\
TimesNet~\cite{wuTimesNetTemporal2DVariation2023} & 0.3375 $\pm$ 0.0027 & 0.6697 $\pm$ 0.0011 & 0.3910 $\pm$ 0.0042 & 0.7103 $\pm$ 0.0012 & 0.4115 $\pm$ 0.0029 \\
PatchTST~\cite{nieTimeSeriesWorth2023} & 0.3360 $\pm$ 0.0013 & 0.6746 $\pm$ 0.0002 & 0.3902 $\pm$ 0.0010 & 0.7167 $\pm$ 0.0002 & 0.4067 $\pm$ 0.0008 \\
xLSTMAD~\cite{faberXLSTMADPowerfulXLSTMbased2025} & 0.3420 $\pm$ 0.0033 & 0.7415 $\pm$ 0.0010 & 0.3895 $\pm$ 0.0037 & 0.7804 $\pm$ 0.0008 & 0.4254 $\pm$ 0.0035 \\
FITS~\cite{xuFITSModelingTime2024} & 0.3053 $\pm$ 0.0004 & 0.6509 $\pm$ 0.0003 & 0.3642 $\pm$ 0.0008 & 0.6994 $\pm$ 0.0001 & 0.3702 $\pm$ 0.0006 \\
MSCRED~\cite{chuxuzhangDeepNeuralNetwork} & 0.3500 $\pm$ 0.0009 & 0.7123 $\pm$ 0.0008 & 0.3622 $\pm$ 0.0018 & 0.7365 $\pm$ 0.0006 & 0.3989 $\pm$ 0.0012 \\
OFA~\cite{zhouOneFitsAll2023} & 0.3005 $\pm$ 0.0033 & 0.6397 $\pm$ 0.0010 & 0.3557 $\pm$ 0.0028 & 0.6887 $\pm$ 0.0008 & 0.3692 $\pm$ 0.0028 \\
MMPAD~\cite{yehMatrixProfileTimeSeries2026} & 0.3091 $\pm$ 0.0000 & 0.7444 $\pm$ 0.0000 & 0.3532 $\pm$ 0.0000 & 0.7582 $\pm$ 0.0000 & 0.3581 $\pm$ 0.0000 \\
MTAD\_GAT~\cite{zhao2020Multivariate} & 0.3648 $\pm$ 0.0007 & 0.7361 $\pm$ 0.0016 & 0.3506 $\pm$ 0.0002 & 0.7608 $\pm$ 0.0013 & 0.4143 $\pm$ 0.0009 \\
GDN~\cite{deng2021Graph} & 0.3106 $\pm$ 0.0036 & 0.7253 $\pm$ 0.0037 & 0.3318 $\pm$ 0.0040 & 0.7578 $\pm$ 0.0029 & 0.3684 $\pm$ 0.0022 \\
OmniAnomaly~\cite{suRobustAnomalyDetection2019} & 0.3152 $\pm$ 0.0000 & 0.7058 $\pm$ 0.0006 & 0.3153 $\pm$ 0.0001 & 0.7435 $\pm$ 0.0006 & 0.3752 $\pm$ 0.0001 \\
CNN~\cite{munirDeepAnTDeepLearning2019} & 0.2934 $\pm$ 0.1016 & 0.6446 $\pm$ 0.2226 & 0.2960 $\pm$ 0.1027 & 0.6728 $\pm$ 0.2330 & 0.3461 $\pm$ 0.1200 \\
CST\_GL~\cite{zhengCorrelationAwareSpatialTemporal2024} & 0.2771 $\pm$ 0.0021 & 0.6787 $\pm$ 0.0022 & 0.2937 $\pm$ 0.0015 & 0.7120 $\pm$ 0.0026 & 0.3321 $\pm$ 0.0023 \\
KMeansAD~\cite{aggarwalOutlierAnalysis2017} & 0.2485 $\pm$ 0.0029 & 0.6869 $\pm$ 0.0074 & 0.2867 $\pm$ 0.0020 & 0.7241 $\pm$ 0.0065 & 0.3052 $\pm$ 0.0024 \\
AE~\cite{kingma2022AutoEncoding} & 0.2635 $\pm$ 0.0050 & 0.6674 $\pm$ 0.0028 & 0.2835 $\pm$ 0.0020 & 0.6951 $\pm$ 0.0015 & 0.3208 $\pm$ 0.0062 \\
PCA~\cite{aggarwalOutlierAnalysis2017} & 0.2409 $\pm$ 0.0000 & 0.6759 $\pm$ 0.0000 & 0.2691 $\pm$ 0.0000 & 0.7058 $\pm$ 0.0000 & 0.3120 $\pm$ 0.0000 \\
TranAD~\cite{tuli2022TranAD} & 0.2654 $\pm$ 0.0922 & 0.5922 $\pm$ 0.2052 & 0.2651 $\pm$ 0.0919 & 0.6214 $\pm$ 0.2152 & 0.3192 $\pm$ 0.1105 \\
OCSVM~\cite{scholkopfEstimatingSupportHighdimensional2001} & 0.2354 $\pm$ 0.0000 & 0.6341 $\pm$ 0.0000 & 0.2628 $\pm$ 0.0000 & 0.6697 $\pm$ 0.0000 & 0.2890 $\pm$ 0.0000 \\
IForest~\cite{liuIsolationForest2008} & 0.2095 $\pm$ 0.0000 & 0.7040 $\pm$ 0.0000 & 0.2487 $\pm$ 0.0000 & 0.7437 $\pm$ 0.0000 & 0.2798 $\pm$ 0.0000 \\
RobustPCA~\cite{xuRobustPCAOutlier2010} & 0.2386 $\pm$ 0.0000 & 0.5898 $\pm$ 0.0002 & 0.2379 $\pm$ 0.0000 & 0.6122 $\pm$ 0.0002 & 0.2833 $\pm$ 0.0000 \\
EIF~\cite{haririExtendedIsolationForest2021} & 0.1850 $\pm$ 0.0016 & 0.6685 $\pm$ 0.0017 & 0.2052 $\pm$ 0.0013 & 0.7057 $\pm$ 0.0018 & 0.2574 $\pm$ 0.0021 \\
COPOD~\cite{liCOPODCopulaBasedOutlier2020} & 0.2046 $\pm$ 0.0000 & 0.6518 $\pm$ 0.0000 & 0.2015 $\pm$ 0.0000 & 0.6820 $\pm$ 0.0000 & 0.2688 $\pm$ 0.0000 \\
HBOS~\cite{dengelHistogrambasedOutlierScore} & 0.1610 $\pm$ 0.0000 & 0.6331 $\pm$ 0.0000 & 0.1856 $\pm$ 0.0000 & 0.6691 $\pm$ 0.0000 & 0.2439 $\pm$ 0.0000 \\
KNN~\cite{ramaswamyEfficientAlgorithmsMining2000} & 0.1317 $\pm$ 0.0000 & 0.4961 $\pm$ 0.0000 & 0.1693 $\pm$ 0.0000 & 0.5667 $\pm$ 0.0000 & 0.1826 $\pm$ 0.0000 \\
LOF~\cite{breunigLOFIdentifyingDensitybased2000} & 0.0953 $\pm$ 0.0000 & 0.5321 $\pm$ 0.0000 & 0.1310 $\pm$ 0.0000 & 0.5831 $\pm$ 0.0000 & 0.1493 $\pm$ 0.0000 \\
AT~\cite{xu2022Anomaly} & 0.0643 $\pm$ 0.0226 & 0.3978 $\pm$ 0.1360 & 0.0989 $\pm$ 0.0351 & 0.4286 $\pm$ 0.1476 & 0.1021 $\pm$ 0.0360 \\
CARMA (Ours) & \textbf{0.4425} $\pm$ 0.0009 & \textbf{0.8330} $\pm$ 0.0017 & \textbf{0.4929} $\pm$ 0.0027 & \textbf{0.8598} $\pm$ 0.0010 & \textbf{0.5301} $\pm$ 0.0032 \\
\bottomrule
\end{tabular}
\end{table}

\subsection{CARMA vs Single Latent (w/o $z_c$) Variant Ablation}
\label{app:abl_carma_vs_nocorr}
\begin{figure}
    \centering
    \begin{minipage}[c]{0.49\linewidth}
        \centering
        \includegraphics[width=\linewidth]{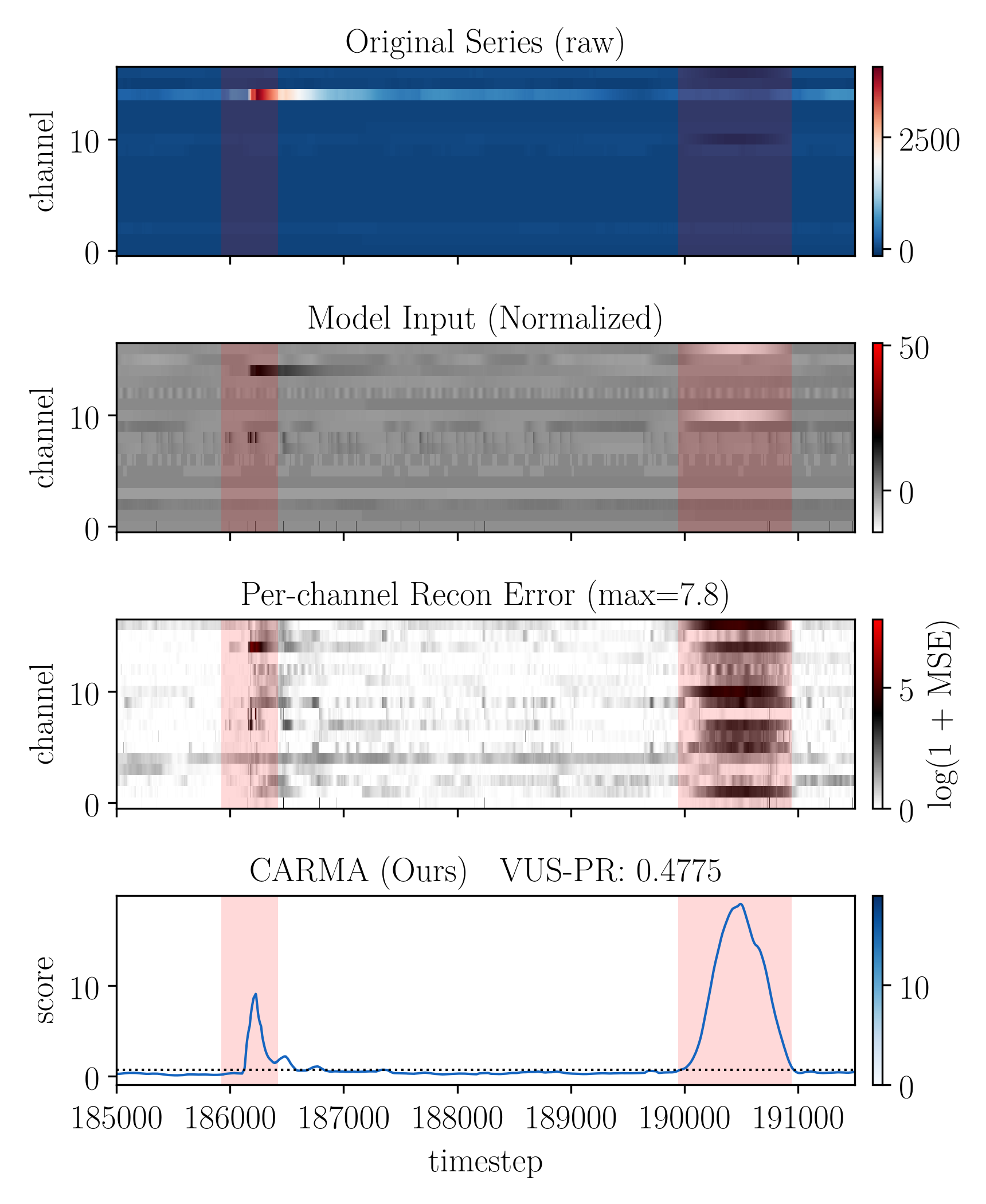}
        \subcaption{CARMA (Ours).}
    \end{minipage}%
    \hfill
    \begin{minipage}[c]{0.49\linewidth}
        \centering
        \includegraphics[width=\linewidth]{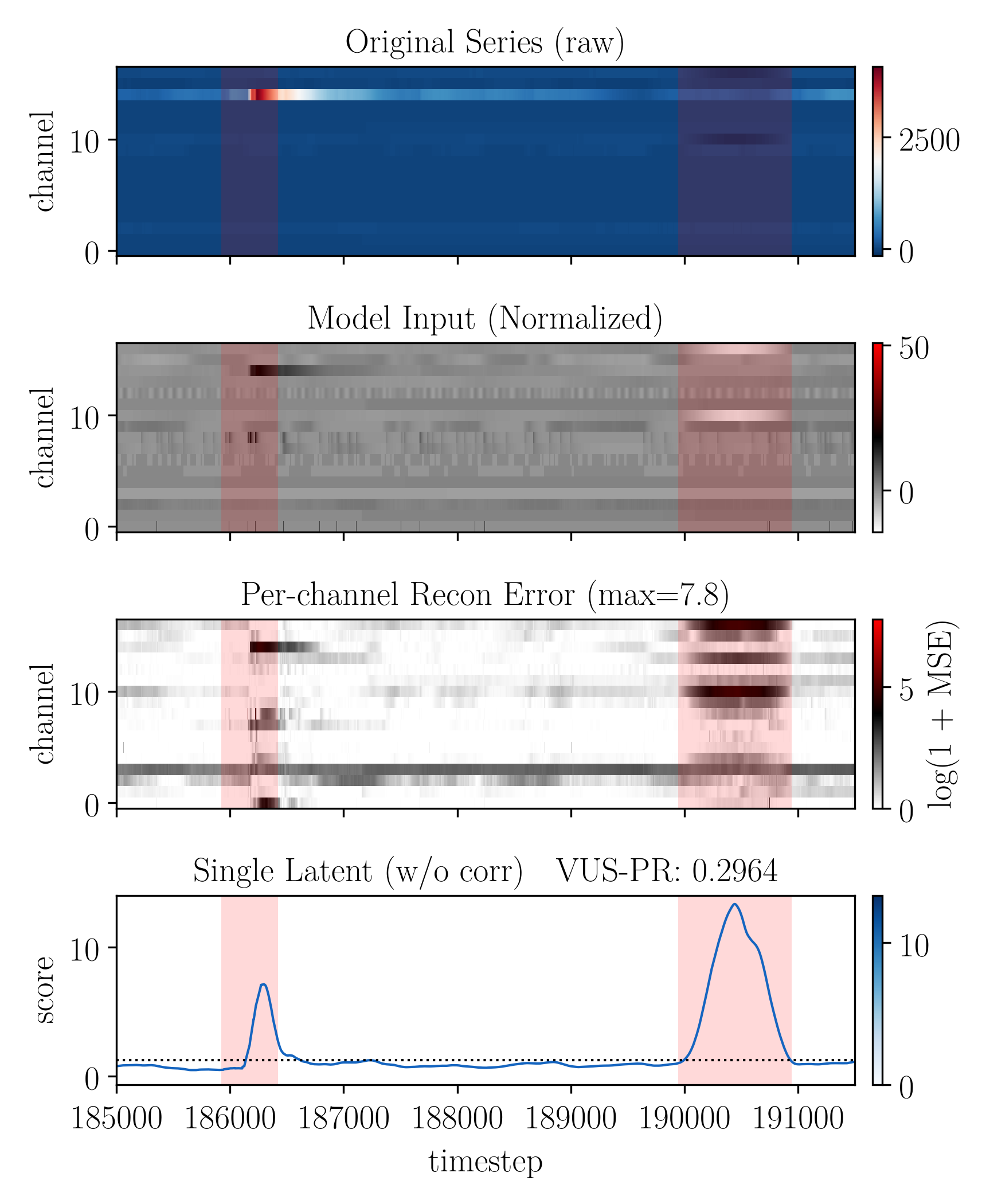}
        \subcaption{Single Latent Variant (w/o $z_c$).}
    \end{minipage}
    \caption{Side-by-side comparison of CARMA and single latnet model (w/o $z_c$) on one TSB-AD-M series. Each column shows the raw multivariate series, the normalized input, per-channel reconstruction error as log(1 + MSE), and anomaly score. Red bands mark ground-truth anomalies. CARMA (Ours) shows a clearer score peak at the event and a higher VUS-PR (0.48 vs 0.30).}
    \label{fig:case_study}
\end{figure}

\textbf{Case Study.} We selected the series 141\_CATSv2\_id\_4\_Sensor\_tr\_41727\_1st\_41827 since it contains inter-channel anomalies in the sense of Fleith et al.~\cite{fleithControlledAnomaliesTime2023}. In the top row of Figure~\ref{fig:case_study}, we illustrate the multivariate time series as a heatmap to highlight cross-channel behaviors during anomalous intervals. The second row is the normalized input passed to the model. The third row is per-channel reconstruction error in $\log(1 + \mathrm{MSE})$. The fourth row is the anomaly score of the two variants. Red vertical bands mark two ground-truth segments, a shorter event near timestep 186{,}000 and a longer event between 190{,}000 and 191{,}000.

Outside those bands the series is largely normal. In the third row the w/o corr panel often shows broader patches of elevated error on normal timesteps, so normal periods are not reconstructed as cleanly. The correlated model keeps the normal portion closer to the background level and raises error more selectively toward the annotated intervals. That pattern carries into the fourth row. On the first anomaly segment both variants increase the score, but with correlation the rise is sharper relative to the surrounding normal region. On the second segment the correlated score forms a clearer sustained peak aligned with the band, whereas w/o corr the same interval mixes with more baseline fluctuation and the peak is less distinct. Overall, correlation improves separation between normal reconstruction and event-driven error on this inter-channel example, which is reflected in the higher VUS-PR on this series (0.48 vs 0.30).

\begin{figure}
    \centering
    \begin{minipage}[b]{0.74\linewidth}
        \centering
        \begin{subfigure}[b]{0.32\linewidth}
            \centering
            \includegraphics[width=\linewidth]{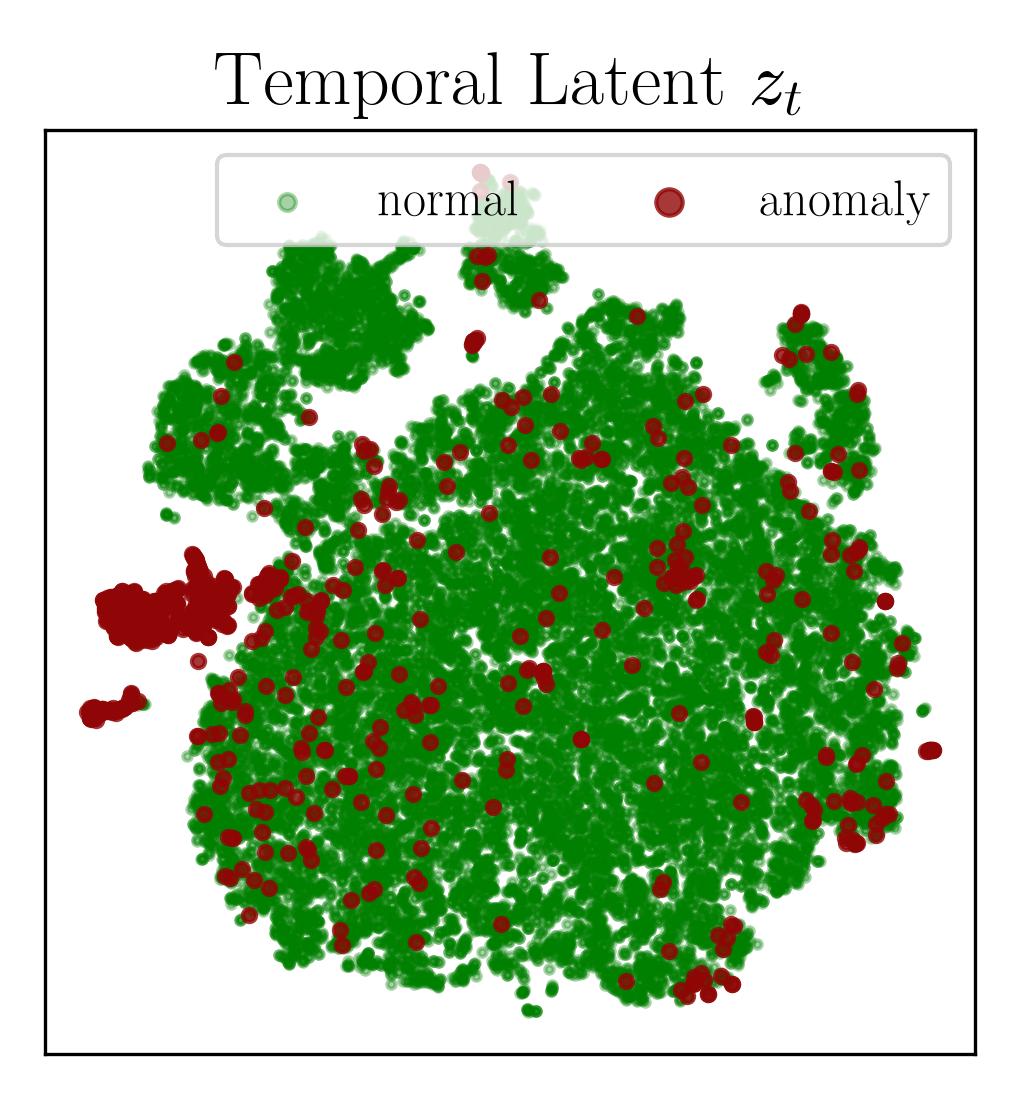}
            \subcaption*{Temporal Latent $z_t$}
        \end{subfigure}\hfill
        \begin{subfigure}[b]{0.32\linewidth}
            \centering
            \includegraphics[width=\linewidth]{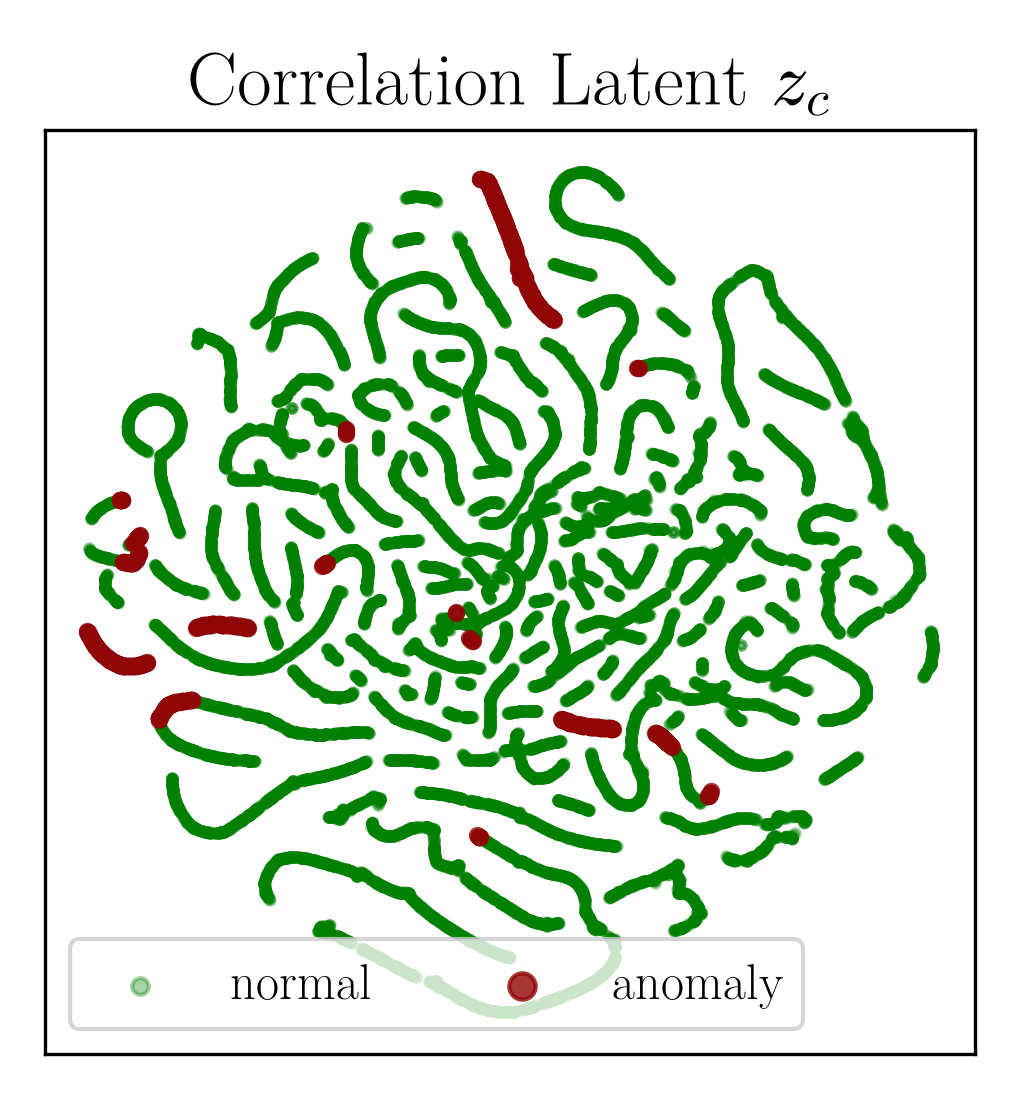}
            \subcaption*{Correlation Latent $z_c$}
        \end{subfigure}\hfill
        \begin{subfigure}[b]{0.32\linewidth}
            \centering
            \includegraphics[width=\linewidth]{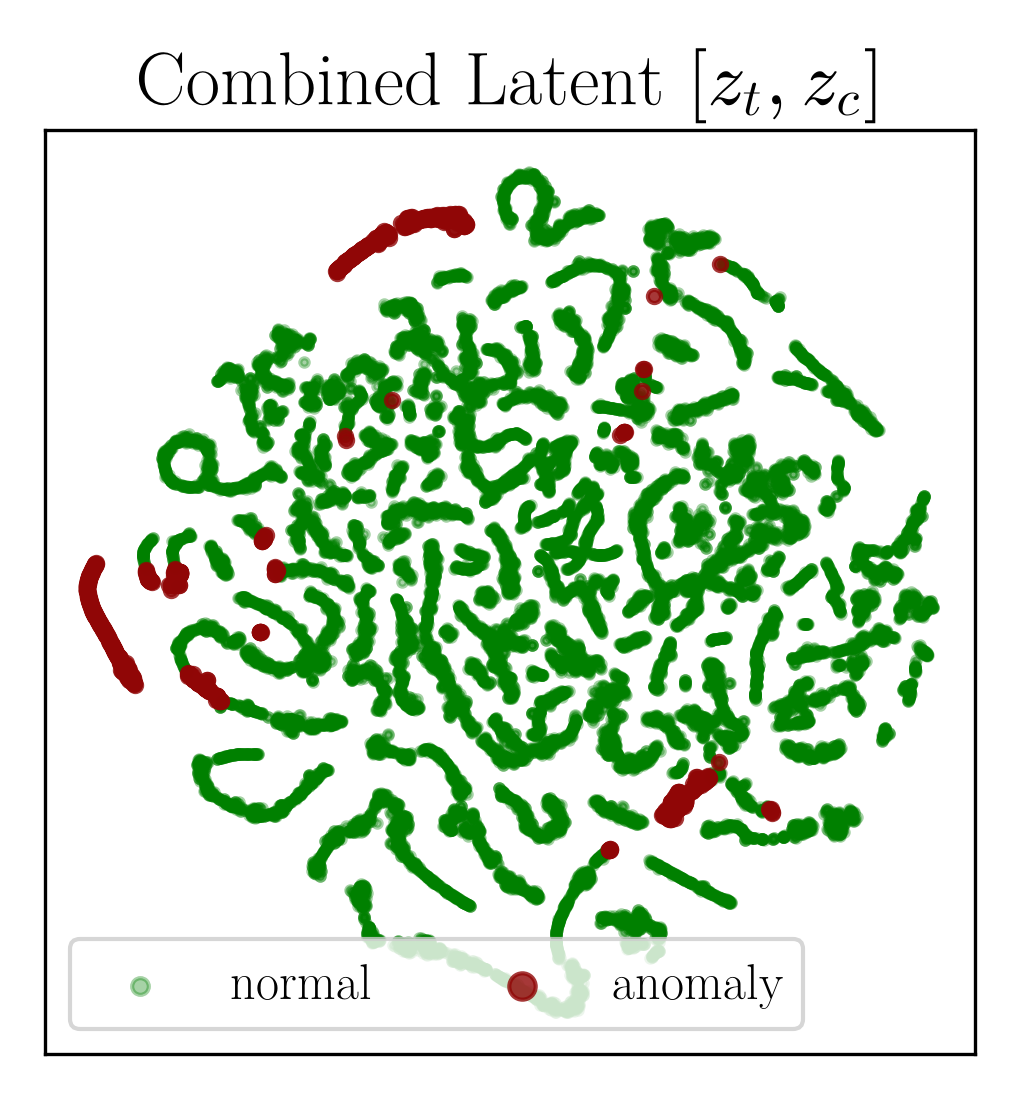}
            \subcaption*{Combined Latent $[z_t, z_c]$}
        \end{subfigure}\\[4pt]
        \subcaption{CARMA (Ours).}
        \label{fig:corr}
    \end{minipage}%
    \hfill
    \begin{minipage}[b]{0.24\linewidth}
        \centering
        \begin{subfigure}[b]{\linewidth}
            \centering
            \includegraphics[width=\linewidth]{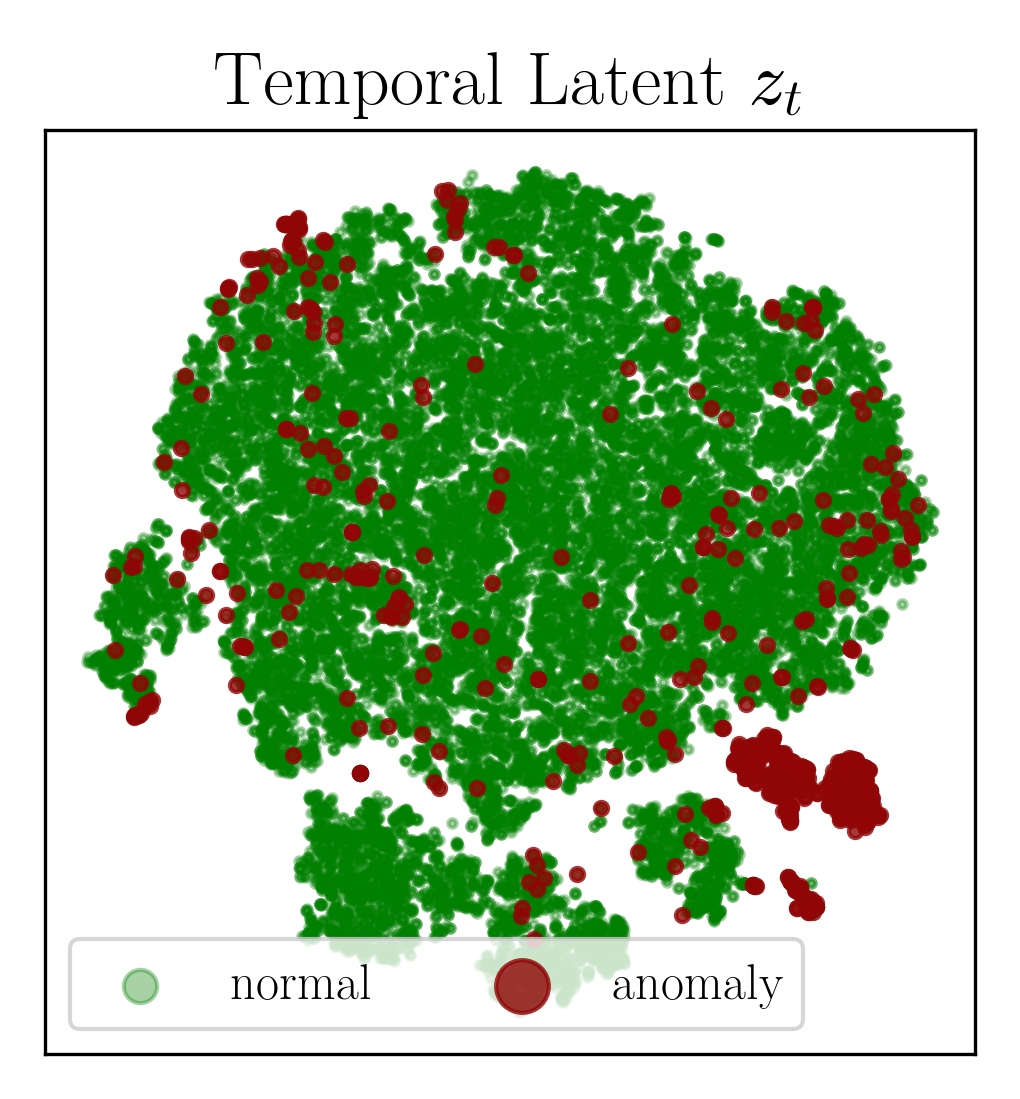}
            \subcaption*{Temporal Latent $z_t$}
        \end{subfigure}\\[4pt]
        \subcaption{\footnotesize Single Latent (w/o $z_c$).}
        \label{fig:nocorr}
    \end{minipage}
\caption{t-SNE visualization of latent representations for series \texttt{SMD\_id\_10\_Facility\_tr\_2634\_1st\_2734}. 
\textbf{(a) CARMA (Ours):} Panels show the per-timestep temporal latent $z_t$, the window-level correlation latent $z_c$, and their concatenation $[z_t, z_c]$. 
\textbf{(b) Singel latent (w/o $z_c$)} The per-timestep temporal latent $z_t$ obtained when the correlation branch is ablated.}
    \label{fig:latent_vis}
\end{figure}

\textbf{Visualizing Embeddings.} Figure~\ref{fig:latent_vis} illustrates the latent space encoding of a single SMD series.  The temporal latent $z_t$ forms a dense cluster where normal and anomalous timesteps significantly overlap, indicating that time-dependent features alone do not sufficiently disentangle faults.  In contrast, the window-level correlation code $z_c$ exhibits a distinct filamentary manifold structure as in the figure anomalies concentrate along specific curved branches rather than scattering through the bulk normal distribution. The concatenated representation $[z_t, z_c]$ preserves this path-like geometry while retaining per-timestep granularity. When the correlation branch is ablated (Single latent (w/o $z_c$)), the model relies solely on $z_t$. Although anomalies form isolated clusters at the periphery, they remain heavily mixed with normal points in the central mass. This suggests that the correlation latent may capture structural information complementary to the temporal branch, aiding in better AD.

\subsection{Synthetic Benchmark}
\label{app:synthetic}
\textbf{Data Generation.} Following the methodology of Pinet et al. \cite{pinet2026Anomalies}, we created a scaled-up variant of the NoiseFlip and NoiseRoll benchmarks to better evaluate the AD methods. We increased the channel count to C=55 and expanded both the training to 20,000 timesteps and test sets to 100,000 timesteps, with anomalies injected into 15\% of the test series. Anomalous segment lengths were constrained to the range [300, 500] timesteps, separated by a minimum inter-segment gap of 50 timesteps and a 500-timestep buffer at the beginning and end of the test series to avoid edge effects. Apart from these scaling and boundary adjustments, the core anomaly injection mechanisms for both NoiseFlip and NoiseRoll remain identical to those described in the original work \cite{pinet2026Anomalies}. 

\subsection{Evaluation Metrics}
\label{app:eval_explain}
This section defines the evaluation metrics used in the main paper. Threshold-free ranking is assessed via AUC-ROC, AUC-PR, VUS-ROC, and VUS-PR. AUC-ROC plots the true positive rate against the false positive rate across all thresholds, while AUC-PR trades precision against recall and is the more informative choice under class imbalance, where it concentrates on the positive class. VUS-ROC \cite{paparrizosVolumeSurfaceNew2022} generalizes AUC-ROC by integrating the score over temporal buffer zones around ground truth events, so that predictions near an event are not penalized as false positives. VUS-PR \cite{boniol2025VUS} applies the same volume-under-surface construction in precision-recall space.  Localization is evaluated using Standard-F1, PA-F1 \cite{kimRigorousEvaluationTimeseries2022}, Event-based-F1, R-based-F1 \cite{tatbulPrecisionRecallTime2019}, and Affiliation-F. Standard-F1 computes precision and recall at a fixed threshold and ignores temporal structure. PA-F1 instead marks an entire predicted segment as correct when it overlaps any ground-truth anomaly. Event-based-F1 matches predicted and true segments by intersection over union, accounting for both onset and offset. Finally, R-based-F1 and Affiliation-F both assign partial credit for predictions near a true anomaly. Statistical significance of performance differences is assessed via a critical difference diagram \cite{demsarStatisticalComparisonsClassifiers2006}. The diagram ranks methods by their average rank across all series and connects groups whose pairwise differences are not statistically significant at a chosen confidence level, so that methods within a connected clique should be treated as statistically indistinguishable.

\end{document}